\documentclass{article}

\PassOptionsToPackage{numbers, compress}{natbib}

\usepackage[final]{neurips_2025}

\usepackage[utf8]{inputenc} 
\usepackage[T1]{fontenc}    
\usepackage{url}            
\usepackage{amsfonts}       
\usepackage{nicefrac}       
\usepackage{microtype}      
\usepackage{xcolor}         
\usepackage{booktabs}       
\usepackage{colortbl}
\usepackage{xspace}
\usepackage{amsmath}
\usepackage{multirow}
\usepackage{graphicx}   
\usepackage{array} 
\usepackage{amssymb}
\usepackage{adjustbox}
\usepackage{multicol}
\usepackage{makecell} 
\usepackage{wrapfig}
\usepackage{booktabs}
\definecolor{citecolor}{HTML}{2980b9}
\definecolor{linkcolor}{HTML}{c0392b}
\definecolor{mycolor_blue}{HTML}{E7EFFA}
\definecolor{mycolor_green}{HTML}{E6F8E0}
\definecolor{mycolor_gray}{HTML}{ECECEC}
\definecolor{pearDark}{HTML}{2980B9}
\definecolor{citecolor}{HTML}{2980b9}
\definecolor{linkcolor}{HTML}{c0392b}
\definecolor{sem}{HTML}{2E75B6}
\definecolor{tok}{HTML}{F3B000}
\definecolor{royalblue}{HTML}{4169E1}
\definecolor{mypurple}{HTML}{7B68EE}
\definecolor{darkorange}{HTML}{FAA460}
\definecolor{dimgray}{HTML}{696969}
\usepackage{enumitem}
\usepackage[hidelinks,breaklinks=true,colorlinks,citecolor=citecolor,linkcolor=linkcolor]{hyperref}
\usepackage{cleveref}
\usepackage{subcaption}
\newcolumntype{C}[1]{>{\centering\arraybackslash}m{#1}}


\newcommand{\eg}{\textit{e.g.},\xspace}

\makeatletter
\providecommand{\@trackname}{} 
\makeatother

\usepackage[most]{tcolorbox}
\usepackage{xcolor}
\usepackage[most]{tcolorbox}

\definecolor{TakeGradTop}{HTML}{F3F3F3}
\definecolor{TakeGradBottom}{HTML}{FCFCFC}
\newcommand{\TakeawayTitle}{Takeaway}

\newcounter{takeaway}

\newtcolorbox{takeawaybox}[1][]{
  enhanced, breakable,
  boxrule=0pt, colframe=white,              
  interior style={top color=mycolor_blue, bottom color=TakeGradBottom},
  arc=2mm,                                  
  left=8pt, right=8pt, bottom=10pt,
  top=22pt,
  before skip=16pt plus 4pt, after skip=10pt,
  overlay={
    \node[anchor=west,font=\bfseries,black]
      at ([xshift=8pt,yshift=-11pt]frame.north west)
      {\TakeawayTitle~\refstepcounter{takeaway}\thetakeaway};
  },
  #1
}

\usepackage[most]{tcolorbox}
\usepackage{xcolor}
\usepackage{enumitem}

\newtcolorbox{graylist}{
  enhanced,
  colback=gray!5,    
  colframe=gray!15, 
  boxrule=0pt,      
  arc=0mm,          
  left=2mm, right=2mm, top=1mm, bottom=1mm,
  width=\dimexpr\linewidth-4mm\relax,  
  center,            
  breakable
}

\usepackage{pifont}     

\newcommand{\good}{\textcolor{green!50!black}{\emph{Good}~}}
\newcommand{\moderate}{\textcolor{orange!80!black}{\emph{Moderate}~}}
\newcommand{\bad}{\textcolor{red!70!black}{\emph{Bad}}}

\newcommand{\chgood}{\textcolor{green!50!black}{\emph{\checkmark~Good}}}
\newcommand{\chmoderate}{\textcolor{orange!80!black}{\emph{\textasciitilde~Moderate}}}
\newcommand{\chbad}{\textcolor{red!70!black}{\emph{\ding{55}~Bad}}}

\title{Are Video Models Ready as Zero-Shot Reasoners?\\[0.2em]
{\Large\textit{An Empirical Study with the \textsc{MME-CoF} Benchmark}}}
\vspace{0.1cm}

\author{Ziyu Guo$^{*\dagger1}$, Xinyan Chen$^{*2}$, Renrui Zhang$^{*\ddagger2}$, Ruichuan An$^{*3}$, Yu Qi$^{*4}$, Dongzhi Jiang$^{2}$\vspace{0.07cm}\\ 
\textbf{Xiangtai Li$^{3}$, Manyuan Zhang$^{2}$, Hongsheng Li$^{2}$, Pheng-Ann Heng$^{1}$\vspace{0.35cm}}\\
CUHK $^1$IMIXR\hspace{0.1cm} \&\hspace{0.1cm} $^2$MMLab \quad
$^3$Peking University \quad
$^4$Northeastern University\vspace{0.3cm}\\
\small $^*$Equal Contribution\hspace{0.4cm} $^\dagger$Project Lead\hspace{0.4cm} $^\ddagger$Corresponding Author\\
\mbox{}\\[0.05cm]
\centerline{Project Page: \url{https://video-cof.github.io}}
}

\begin{document}
\maketitle

\begin{abstract}
Recent video generation models can produce high-fidelity, temporally coherent videos, indicating that they may encode substantial world knowledge. 
Beyond realistic synthesis, they also exhibit emerging behaviors indicative of visual perception, modeling, and manipulation~\cite{wiedemer2025video}. 
Yet, an important question still remains: \textit{Are video models ready to serve as zero-shot reasoners in challenging visual reasoning scenarios?}
In this work, we conduct \textbf{an empirical study} to comprehensively investigate this question, focusing on the leading and popular Veo-3~\cite{GoogleDeepMind2025Veo3}. 
We evaluate its reasoning behavior across 12 dimensions, including spatial, geometric, physical, temporal, and embodied logic, systematically characterizing both its strengths and failure modes.
To standardize this study, we curate the evaluation data into \textbf{\textsc{MME-CoF}}, a compact benchmark that enables in-depth and thorough assessment of Chain-of-Frame (CoF) reasoning. 
Our findings reveal that while current video models demonstrate promising reasoning patterns on short-horizon spatial coherence, fine-grained grounding, and locally consistent dynamics, they remain limited in long-horizon causal reasoning, strict geometric constraints, and abstract logic. 
Overall, they are \textbf{\textit{not yet reliable}} as standalone zero-shot reasoners, but exhibit encouraging signs as complementary visual engines alongside dedicated reasoning models.
\end{abstract}

\section{Introduction}
\begin{figure}
    \centering
    \includegraphics[width=\linewidth]{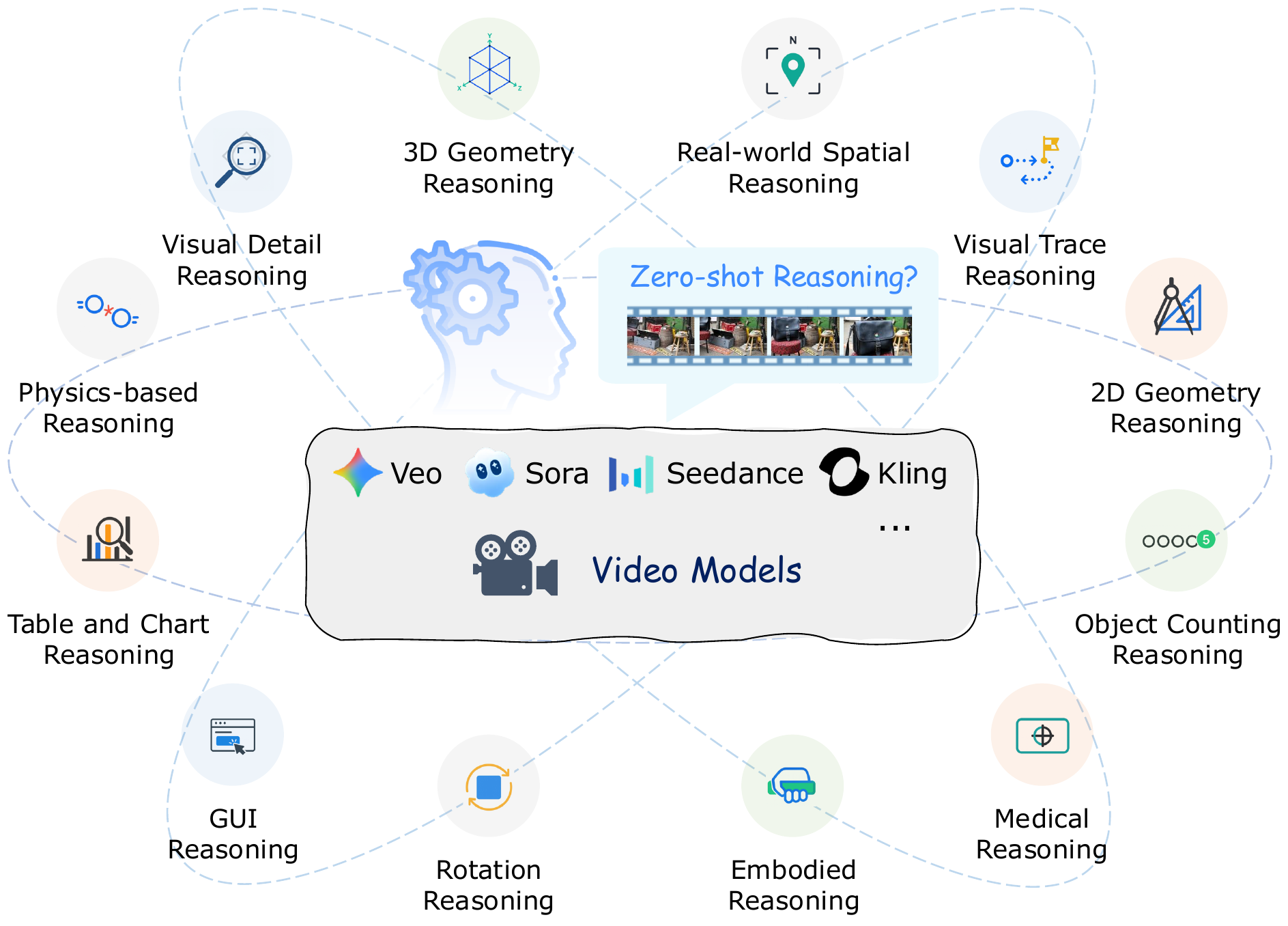}
    \vspace{0.15cm}
    \caption{\textbf{Overview of Our Study on the Reasoning Potential of Video Models.} We investigate whether state-of-the-art video models exhibit emergent reasoning potentials beyond content synthesis. 
    The analysis spans 12 reasoning dimensions under a unified perspective, exploring whether large-scale video models can serve as zero-shot visual reasoners via CoF reasoning.
    }
    \label{fig:intro}
\end{figure}
Video models~\cite{GoogleDeepMind2025Veo3,wan2025wan,openai2024sora,zhang2024llavanext-video,cheng2024videollama}, including text-to-video and video-to-text generation models, have made rapid progress in recent years. 
Thanks to advances in diffusion~\cite{yang2024cogvideox,blattmann2023align,zheng2024open} and autoregressive~\cite{kondratyuk2023videopoet, yu2023magvit,feng2025unified} architectures, current video models can produce high-fidelity videos maintaining consistent object relations and realistic motion dynamics across frames. 
This suggests that the models may have internalized substantial visual and structural knowledge about the world. 
Recent research from Google~\cite{wiedemer2025video} further hints that, such models are evolving beyond pure content generation: Veo-3~\cite{GoogleDeepMind2025Veo3} has been shown to perform dozens of distinct vision tasks across perception, modeling, manipulation, and reasoning, \textit{without} any task-specific training. These emergent capabilities have led researchers to posit that video models could serve as unified, generalist vision models, much like large language models (LLMs)~\cite{achiam2023gpt,dubey2024llama,bai2023qwen,jiang2023mistral7b} have become foundation models for natural language.

Crucially, the sequential nature of video generation provides a new perspective on how such models might reason.
Each generated frame builds upon the last, creating a temporal chain of information propagation. This has been dubbed ``Chain-of-Frame'' (CoF) reasoning~\cite{wiedemer2025video}, an analogy to the chain-of-thought (CoT) process in LLMs~\cite{wei2022chain,kojima2022large,zhang2022automatic,guo2025deepseek,zhang2024mathverse} and their multi-modal variants (MLLMs)~\cite{comanici2025gemini,bai2023qwenvlversatilevisionlanguagemodel,li2024llava,jiang2025mme,chen2025mint}. 
In essence, as a video model generates a sequence of frames, it can iteratively refine and update the scene, thereby working through a problem step-by-step in time and space. 
This CoF concept suggests that, beyond surface-level pattern generation, general-purpose visual reasoning may emerge from video generative models.

However, it remains unclear \textit{to what extent current video models truly exhibit reasoning about the content they create}. 
Strong generative performance does not automatically imply robust reasoning potential. 
Emerging evidence~\cite{guan2025etva, liu2023fetv, bai2025impossible,zhang2025morpheus} shows that a model may produce coherent videos by learning surface-level patterns in the training data, rather than by internalizing general principles. 
For instance, a video model can maintain object continuity yet fail to grasp physical plausibility across a long sequence, or it may mimic observed visual sequences without understanding the underlying cause-and-effect relationships. 
This motivates our central question:  \textbf{\textit{Are video models, purely through large-scale visual learning, obtain the zero-shot reasoning potential?}}

To this end, we present \textbf{the first empirical study} to systematically probe the CoF reasoning capabilities of modern video models, spanning 12 dimensions such as spatial, geometric, physical, temporal, and embodied logic, as detailed in~\ref{fig:intro}. 
We carry out our analysis on Veo-3, which has been systematically examined as a zero-shot learner in prior work~\cite{wiedemer2025video}. 
Our preliminary observations suggest that current leading video models exhibit comparable reasoning patterns, making Veo-3 a representative choice. 
Our analysis builds on reasoning scenarios distilled from diverse reasoning‐oriented benchmarks~\cite{guo2025rbench,wang2025mmbench,li2023super,wu2024v,jia2025omnispatial,kim2024tablevqa}, as well as those we design ourselves, providing a compact yet expressive foundation. 
The prompts for video models are meticulously crafted by transforming the underlying, textual reasoning process of problem-solving into a clear, video-presentation format.
Each case receives a qualitative assessment across three performance levels, i.e., good, moderate, and bad, complemented by a quantitative success rate to measure robustness.

\begin{figure*}[t]
\centering
\begin{minipage}[t]{0.5\textwidth}
    \centering
    \includegraphics[width=\linewidth]{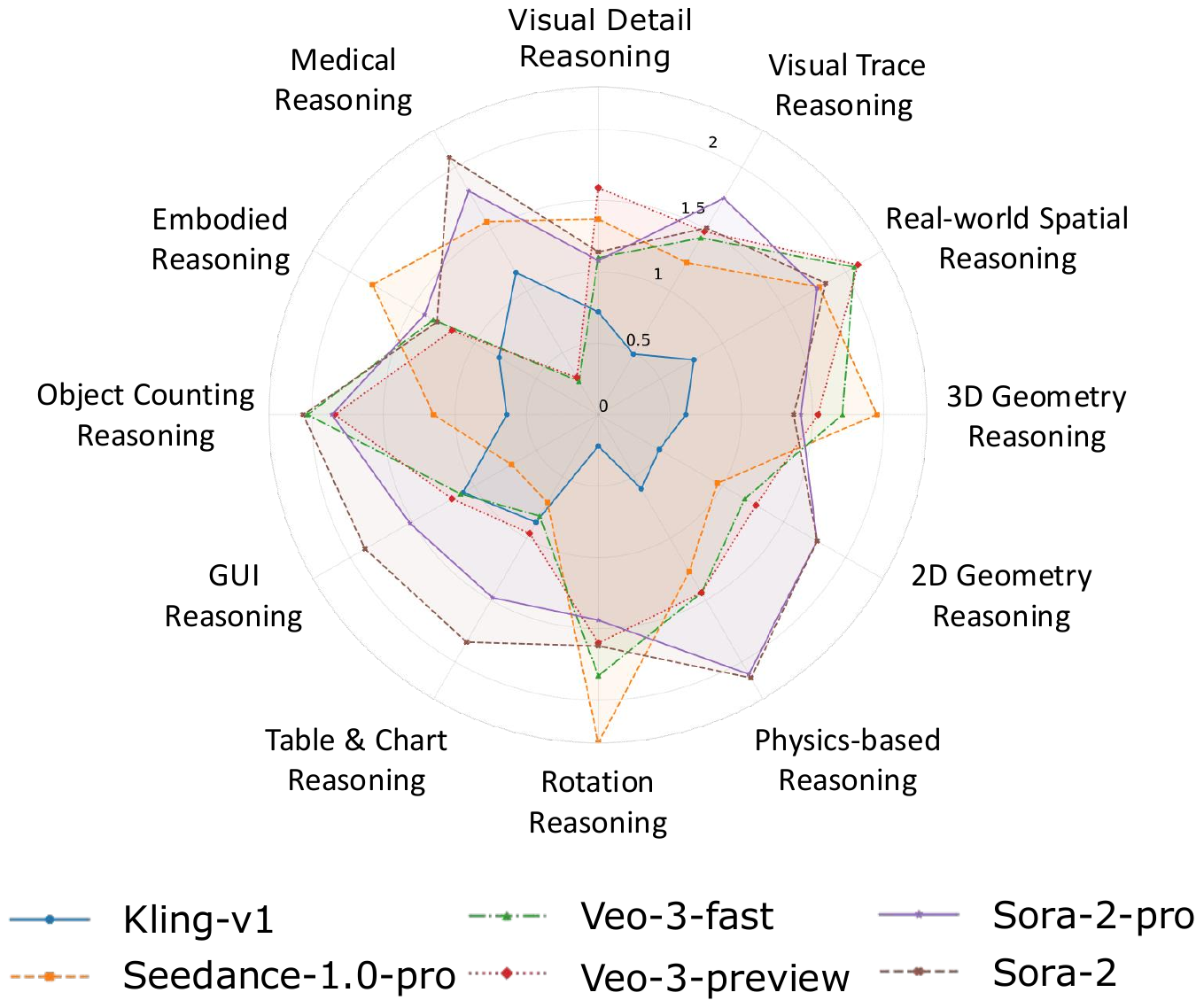}
    \vfill
    \vspace{0.1cm}
    \subcaption{Evaluation Radar Map.}
    \label{fig:overview:radar}
\end{minipage}
\hfill
\begin{minipage}[t]{0.42\textwidth}
    \centering
    \includegraphics[width=\linewidth]{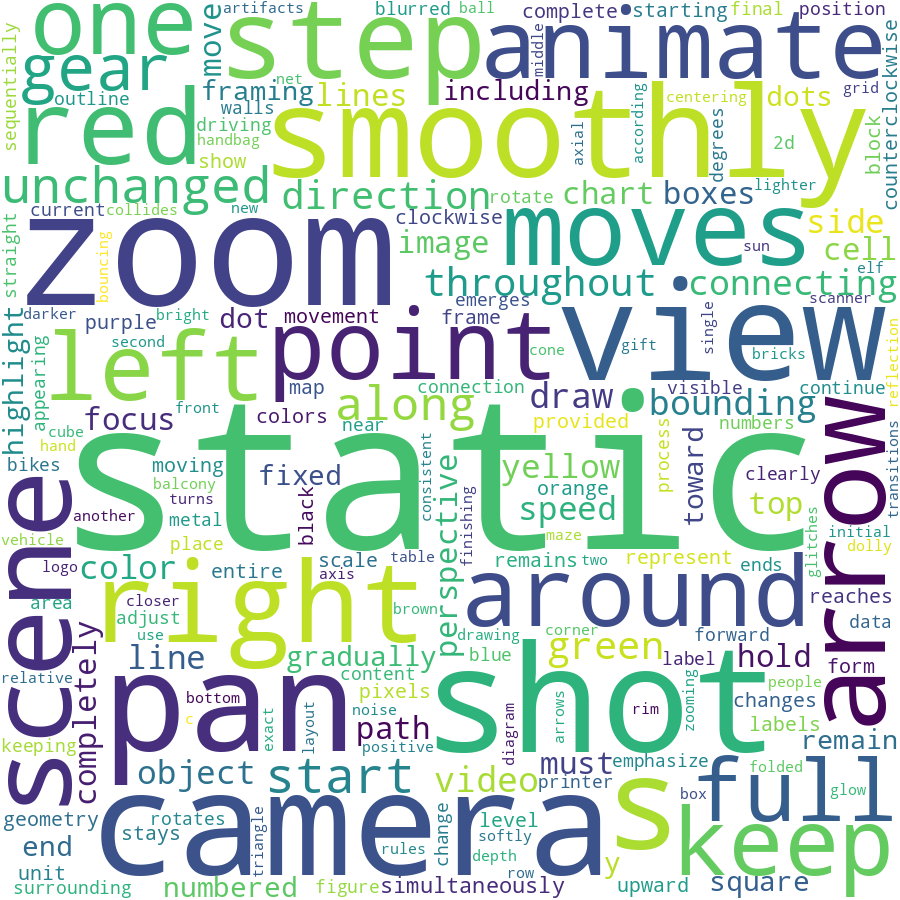}
    \vfill
    \vspace{0.1cm}
    \subcaption{Word Cloud.}
    \label{fig:overview:wc}
\end{minipage}
    \vspace{0.2cm}
\caption{\textbf{Illustration of the \textsc{MME-CoF} Benchmark.} It showcases that different models specialize in distinct reasoning aspects, but most models exhibit limited reasoning capability across all tasks.}
\label{fig:mme_cof_overview1}
\end{figure*}

To standardize evaluation, we curate these tasks into the \textbf{\textsc{MME-CoF}} benchmark, as illustrated in ~\Cref{fig:mme_cof_overview1} and ~\Cref{fig:mme_cof_overview2}. Leveraging this benchmark, we measure several state-of-the-art video models, i.e., Veo-3~\cite{GoogleDeepMind2025Veo3}, Sora-2~\cite{openai2025sora2}, Kling~\cite{kuaishou2024kling}, and Seedance~\cite{gao2025seedance}, to obtain directly comparable scores and qualitative behaviors across categories. 
Our investigation reveals that the models exhibit promising reasoning patterns in short-horizon spatial coherence, fine-grained grounding, and consistent local dynamics; however, they struggle with complex reasoning conditions, particularly in long-horizon causal consistency, geometric constraint adherence, and abstract logic. 
Overall, current video models are \textit{\textbf{not yet ready as standalone zero-shot reasoners}}. 
Still, they show encouraging signs of emergent reasoning, suggesting strong potential as complementary reasoning agents alongside specialized models.

Our main contributions are summarized as follows:

\begin{itemize}

\item \textbf{\textit{A Comprehensive Empirical Study.}} We provide \textit{the first} investigation of video models (Veo-3) to analyze their visual reasoning potential, detailing representative successes, characteristic errors, and the conditions under which CoF reasoning emerges, holds, or breaks.

\item \textbf{\textit{The \textsc{MME-CoF} Benchmark.}} We curate \textsc{MME-CoF}, a compact benchmark providing a standardized taxonomy and an evaluation protocol aligned with CoF reasoning, enabling consistent and category-wise assessment beyond surface-level visual fidelity.

\item \textbf{\textit{Insights and Directions.}} We summarize common \textit{success patterns} (\eg short-horizon coherence and stable spatial layout) and \textit{failure patterns} (\eg long-horizon degradation, violations of basic geometry/physics, and temporal logic), making clear when the behavior reflects genuine reasoning versus pattern replay. 
\end{itemize}

\section{Deep-Dive Analysis on Veo-3}
\label{experiment_and_analysis}

\subsection{Overview}
\label{sec:analysis_overview}
To ensure a rigorous empirical study, we detail our core methodology in this section, including the taxonomy of reasoning tasks, test case curation process, the standardized style for prompt design, and the analysis setup.

\paragraph{Task Taxonomy.} 
To capture different dimensions of reasoning, our study starts from dozens of reasoning-oriented tasks, which can be organized into the following 12 categories:

\begin{center}
\begin{graylist}
\begin{multicols}{2}
\begin{enumerate}[
    label=\textit{\arabic*)},
    leftmargin=1.5em,
    rightmargin=0em,
    itemsep=3pt, topsep=2pt
]
    \item \textit{Visual Detail Reasoning}
    \item \textit{Visual Trace Reasoning}
    \item \textit{Real-world Spatial Reasoning}
    \item \textit{3D Geometry Reasoning}
    \item \textit{2D Geometry Reasoning}
    \item \textit{Physics-based Reasoning}
    \item \textit{Rotation Reasoning}
    \item \textit{Table and Chart Reasoning}
    \item \textit{Object Counting Reasoning}
    \item \textit{GUI Reasoning}
    \item \textit{Embodied Reasoning}
    \item \textit{Medical Reasoning}
\end{enumerate}
\end{multicols}
\end{graylist}
\end{center}

Each category comprises several representative cases selected to test specific aspects of reasoning.

\paragraph{Test Case Curation.}
We recruit five PhD-level experts with deep expertise in text-image reasoning, who are tasked with selecting representative cases from benchmarks~\cite{guo2025rbench,wang2025mmbench,li2023super,masry-etal-2022-chartqa,yang2025mmsi} corresponding to each task category. For each reasoning case, the experts manually constructed text prompts that explicitly or unambiguously define the target reasoning objective, aiming to evaluate the potential of video models for multi-modal reasoning. 

\paragraph{Prompt Design Style.}
\label{prompt_design}
To ensure consistency and fairness, all prompts follow a unified style emphasizing explicit visual constraints, controlled motion, and minimal linguistic ambiguity. Prompts are encouraged to be written in imperative form and designed to reduce variance from language interpretation, focusing the model's behavior on the intended visual reasoning objective. The overall design principles are as follows:

\begin{center}
\begin{graylist}
\begin{enumerate}[
    label=\textit{\arabic*)},
    leftmargin=1.5em,       
    rightmargin=0em,        
    itemsep=2pt, topsep=2pt 
]
    \item \textit{Static camera and fixed viewpoint, unless motion is explicitly required by the task.}
    \item \textit{Stable spatial composition, consistent framing, and unchanging scene layout across frames.}
    \item \textit{Clear specification of allowed and disallowed changes (\eg ``no zoom, no pan, no dolly'') to constrain camera dynamics.}
    \item \textit{Explicit temporal phrasing to control the pace of motion, using cues such as ``instantly'', ``smoothly'', or ``step-by-step''.}
    \item \textit{Avoidance of direct textual hints toward the answer; instructions are purely visual and task-oriented.}
    \item \textit{Inclusion of realistic phrasing and scene context to align with the model's natural video priors while minimizing artifacts.}
\end{enumerate}
\end{graylist}
\end{center}

The standardized prompt style ensures that differences in output primarily reflect the model's internal reasoning potential rather than prompt variability.

\begin{figure}
    \centering
    \includegraphics[width=\linewidth]{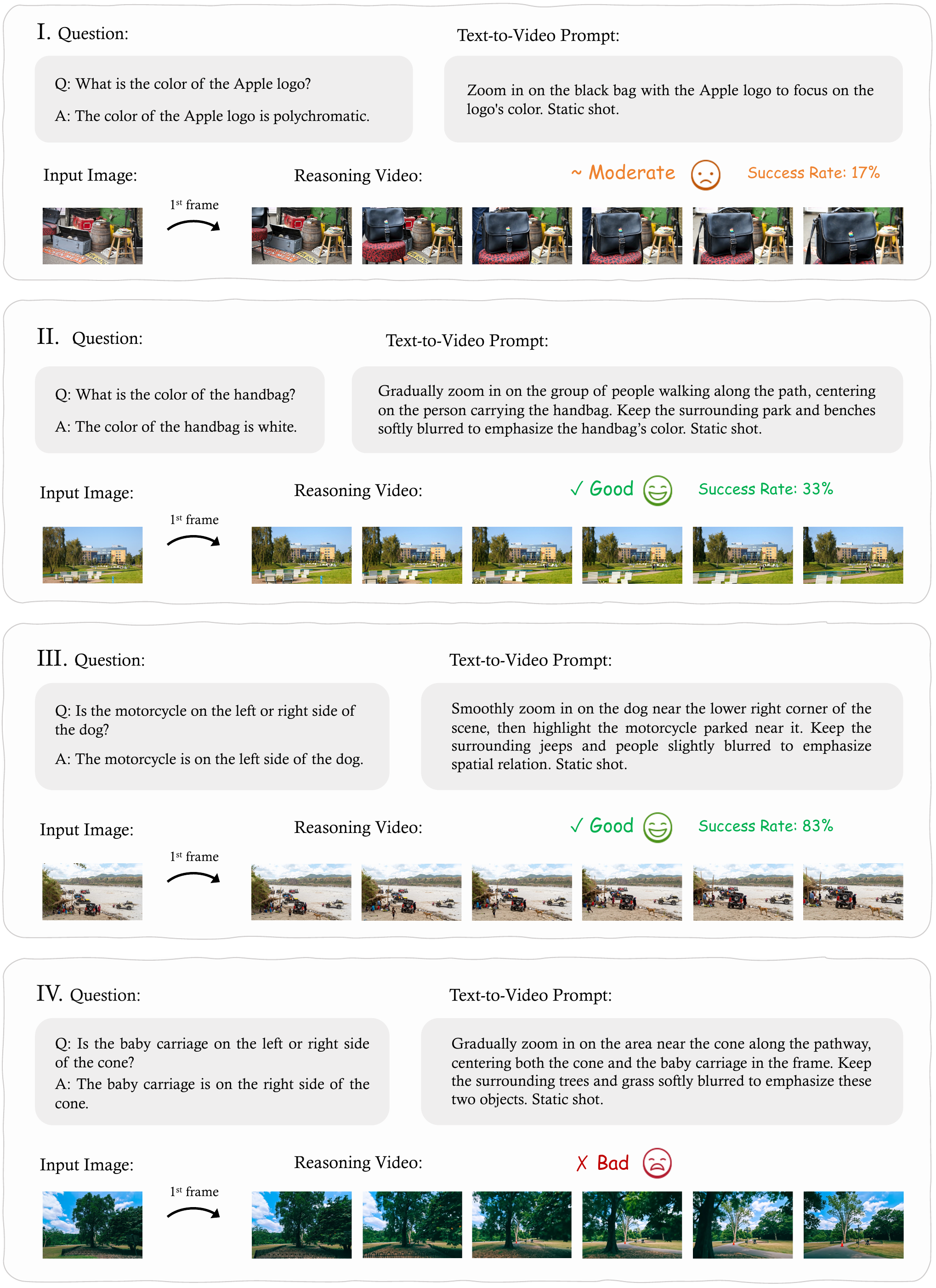}
    \caption{\textbf{Showcase of Visual Detail Reasoning by Veo-3.} It illustrates Veo-3's ability to localize targets and maintain fine-grained visual attributes across frames, together with common failure modes when targets are small, occluded, or embedded in clutter.}
    \label{fig:visual_detail}
\end{figure}

\paragraph{Analysis Setup.}
For every reasoning case, we construct a text prompt that explicitly or implicitly specifies the target reasoning objective. Each prompt produces six video samples at a resolution of 1280$\times$720, 24~FPS, and a duration of 8~seconds. All experiments are conducted in a unified zero-shot setup without fine-tuning, additional supervision, or auxiliary tools. 

We evaluate model outputs through qualitative judgments along three levels of performance, i.e., \good, \moderate, and \bad, based on the clarity, correctness, and temporal stability of the visual reasoning process. 
Detailed definitions and examples of these evaluation criteria are provided in the corresponding task subsections. 
Note that, since we observe that most video models struggle to follow the requirement of `static shot' reliably, we apply more permissive qualitative criteria for static-shot evaluations. 
We further define a \textit{success rate} to measure robustness across generations for each case, computed as the proportion of successful samples among the six generated. 
For cases categorized as \bad, the success rate is always 0. Non-zero success rates only appear in cases evaluated as \good or \moderate, indicating that Veo-3 exhibits some potential to perform as a visual reasoner. 
A higher success rate reflects a more stable reasoning capability of the model.

\subsection{Visual Detail Reasoning}
\paragraph{Task Description and Evaluated Aspects.} In the visual detail reasoning category, the objective is to assess a model's ability to discern and maintain fine‐grained visual attributes and spatial relations within generated video sequences. It covers attribute recognition, \eg identifying color, texture or material of an object, and spatial relation identification, \eg recognizing that one object is on the left of or behind another object. The model is evaluated on the capacity both to attend to the correct target region and to maintain visual consistency, across frames, of the attribute or relation in question.

\paragraph{Definition of \good / \moderate / \bad.}
We define the three-level evaluation criteria as follows:
\begin{center}
\begin{itemize}[
    label={},
    leftmargin=1.5em,
    rightmargin=1.5em,
    itemsep=3pt, topsep=2pt
]
    \item \chgood: The reasoning video accurately centers on the correct target region, clearly resolves the relevant attribute, such as color, texture or position, and maintains sharp, stable and natural rendering throughout the sequence. There are no visible frame drops, artifacts or unintended motion.
    \item \chmoderate: The region of interest is approximately correct, and the attribute remains inferable, but the sequence suffers from minor blur, incomplete framing, slight instability mild unnatural motion, or sometimes deviates from the textual instruction and produces a plausible but unaligned or self-directed visual interpretation, limiting confident interpretation.
    \item \chbad: The target region is incorrect or ambiguous, the attribute cannot be reliably inferred, or the video exhibits severe artifacts: abrupt frame jumps, major jitter, unintended zoom or crop, extraneous objects interfering, or conspicuous quality degradation that obstructs the reasoning task altogether.
\end{itemize}
\end{center}

\paragraph{Data Source.} We sample data from the \textit{V$^*$Bench}~\cite{wu2024v}, which provides a comprehensive set of evaluation dimensions including spatial relationship and color/attribute consistency tasks.

\paragraph{Example and Analysis.}
We illustrate typical behaviors of Veo-3 in visual detail reasoning through four representative cases in ~\Cref{fig:visual_detail}. 
In case I, the model performs well in localizing the target: although it does not strictly execute the ``zoom in'' instruction, it instead achieves an equivalent visual outcome through a semantically consistent motion with a person's hand. 
This slight deviation suggests that the model may exhibit certain generation preferences in how it interprets and realizes spatial instructions, possibly reflecting stylistic tendencies learned from training data. 
In cases II and III, the model achieves better success rates when the targets are visually salient and contextually distinct. 
For the handbag and dog-motorcycle scenes, Veo-3 attends to the correct regions and maintains smooth temporal coherence. 
However, when the object (\eg the motorcycle) is small or surrounded by distracting elements, the model occasionally fails to locate it accurately, indicating limited fine-grained spatial discrimination in cluttered scenes. 
In case IV, when the target object is tiny and visually indistinct, Veo-3 cannot identify it even with explicit positional hints, highlighting that the model's perceptual grounding and reasoning weaken sharply when object size and salience are too low for reliable attention.

\begin{takeawaybox}
Veo-3 performs well in fine-grained attribute and spatial reasoning for salient, well-grounded targets, but fails when objects are small, occluded, or cluttered. It sometimes exhibits stylistic generation biases that lead to plausible yet instruction-divergent outcomes.
\end{takeawaybox}

\subsection{Visual Trace Reasoning}

\begin{figure}
    \centering
    \includegraphics[width=\linewidth]{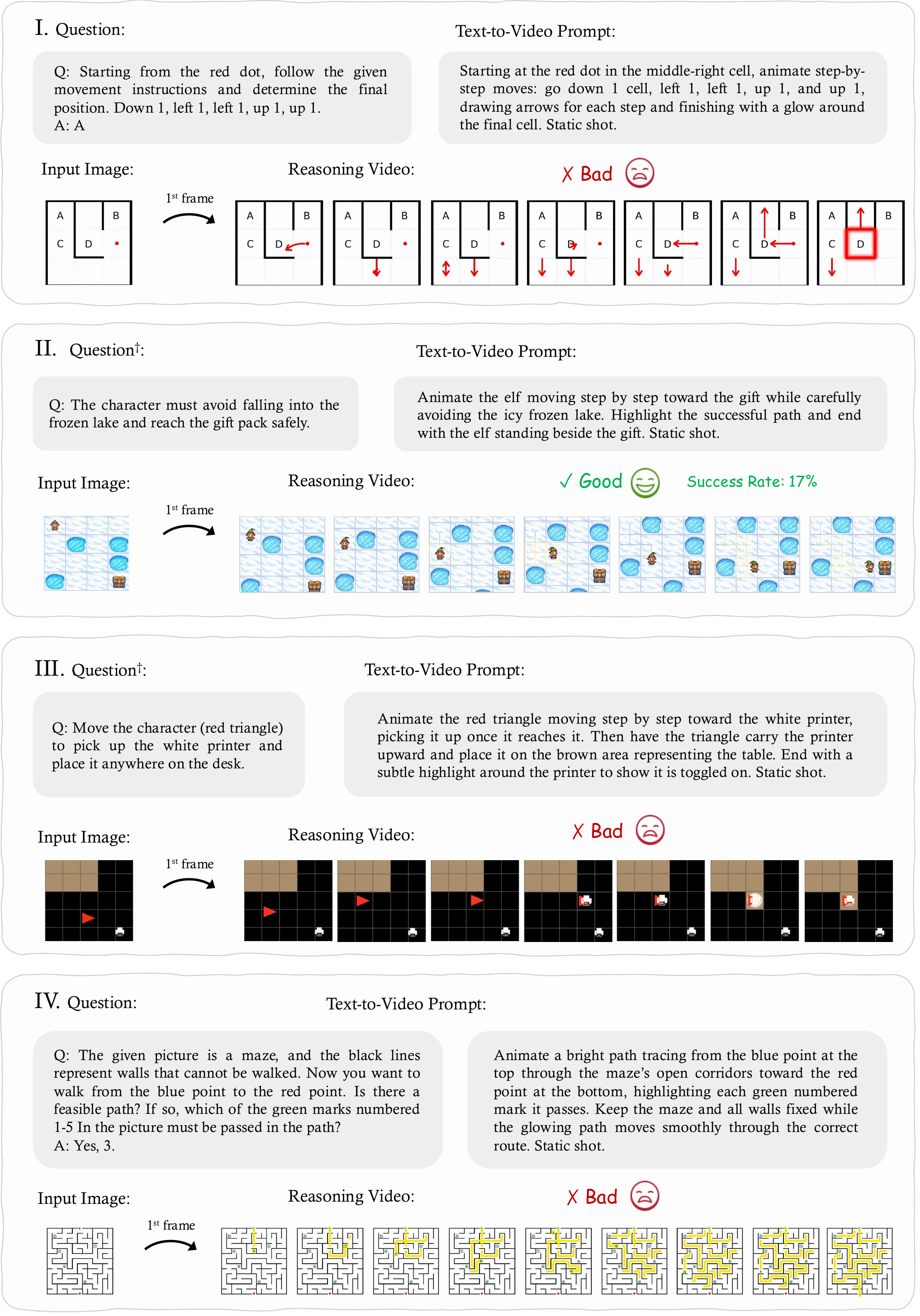}
    \caption{\textbf{Showcase of Visual Trace Reasoning by Veo-3 (Part I).} It shows short-horizon path-following successes, object-grounding failures, and a certain bias that causes step omissions/mistakes in multi-step traces. $^\dagger$ The ground-truth answers of cases II and III are intuitive and non-unique, which are omitted to highlight the key reasoning behaviors.}
    \label{fig:tracing1}
\end{figure}
\begin{figure}
    \centering
    \includegraphics[width=\linewidth]{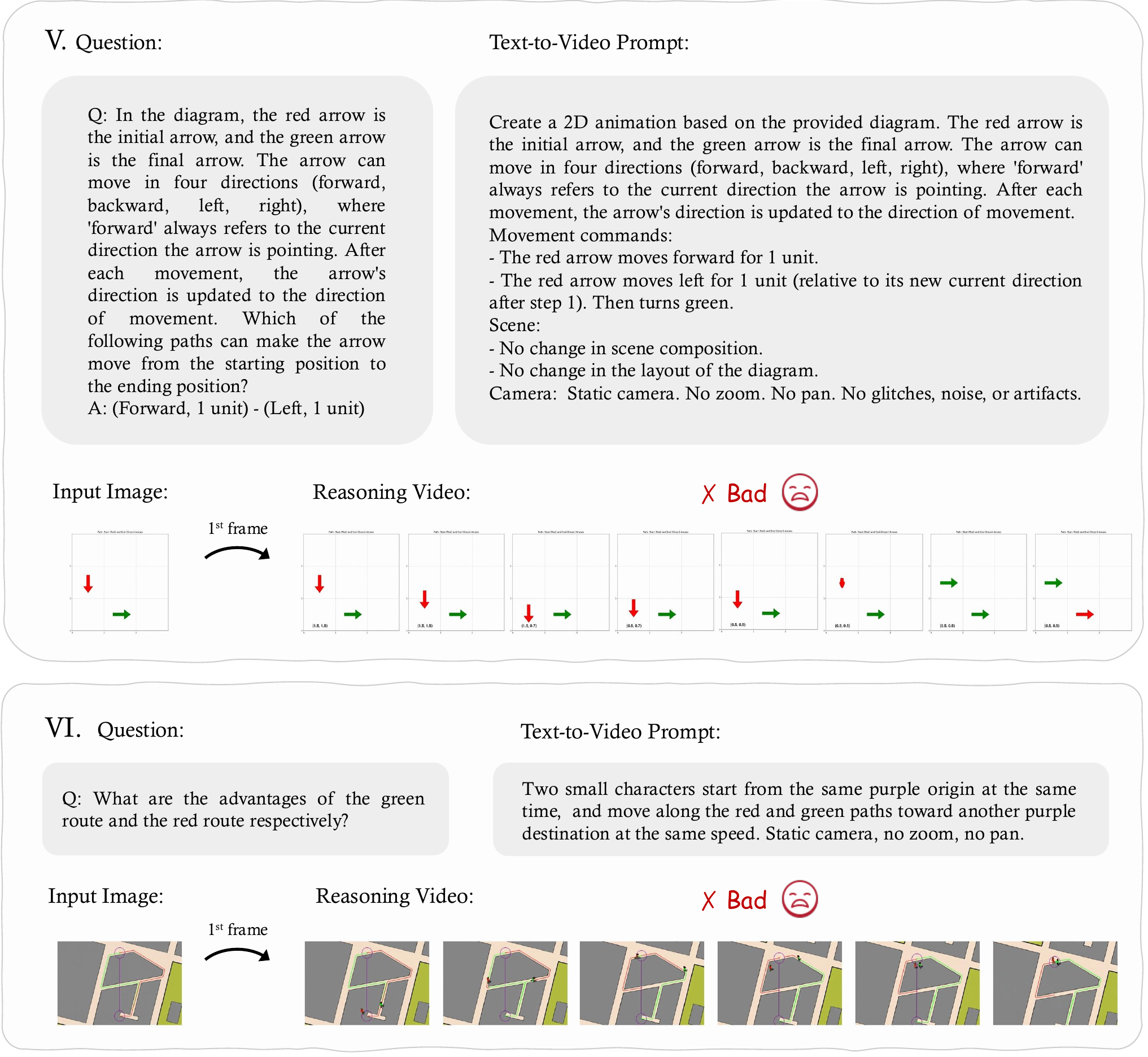}
    \caption{\textbf{Showcase of Visual Trace Reasoning (Part II) by Veo-3.} The examples highlight long-horizon planning breakdowns, inconsistent arrow/trajectory rendering, and failures to preserve comparative or sequential information across frames.}
    \label{fig:tracing2}
\end{figure}

\paragraph{Task Description and Evaluated Aspects.}
The visual trace reasoning category evaluates a model's ability to represent and maintain causal continuity across sequential actions. 
Typical tasks include maze navigation, path following, and multi-step object manipulation, where the video must visually encode a coherent sequence of intermediate decisions that lead to the correct goal. 
Performance is assessed based on two major aspects: \textit{(i)}~temporal coherence, which is the smoothness and logical progression between consecutive steps; and \textit{(ii)}~goal consistency, which means whether the full sequence visually completes the intended reasoning trajectory without deviation or contradiction.

\paragraph{Definition of \good / \moderate / \bad.}
We rate the performance according to the following criteria:
\begin{center}
\begin{itemize}[
    label={},
    leftmargin=1.5em,
    rightmargin=1.5em,
    itemsep=3pt, topsep=2pt
]
    \item \chgood: Each movement step is depicted continuously and logically toward the correct goal. The motion is smooth, temporally consistent, and follows causal order with no skipping, stuttering, or direction reversal.
    \item \chmoderate: The overall trajectory roughly aligns with the intended sequence, but small discontinuities, timing irregularities, or partial missteps occur. The reasoning path remains interpretable, and the goal can still be inferred.
    \item \chbad: Key steps are missing, reversed, or illogical. The sequence shows abrupt jumps, inconsistent object trajectories, or goal confusion, breaking the temporal and causal coherence of the reasoning process.
\end{itemize}
\end{center}

\paragraph{Data Source.}
We select samples from \textit{MVoT}~\cite{li2025imagine}, \textit{FrozenLake}~\cite{brockman2016openai,wu2024vsp}, \textit{MiniBehavior}~\cite{jin2023mini}, \textit{RBench-V}~\cite{guo2025rbench}, \textit{SpatialViz-Bench}~\cite{wang2025spatialviz}, and \textit{OmniSpatial}~\cite{jia2025omnispatial}, which provide controlled multi-step environments for evaluating temporal reasoning, sequential planning, and causal continuity in visual simulations.

\paragraph{Example and Analysis.} 
In ~\Cref{fig:tracing1} and ~\Cref{fig:tracing2}, we showcase six representative visual-trace examples. In case I, the model repeatedly fails to execute the exact step sequence and instead drifts toward a visually salient central cell. However, case II is one of the few successes: the model can produce a coherent step-by-step path in simple, low-branching settings, but this behavior is not robust across trials. Case III largely fails, where the model often does not ground the specified object (printer), sometimes hallucinating its appearance or placement rather than performing a consistent pickup-and-place. Case IV shows near-uniform failure on long-horizon, highly branched navigation: outputs contain wrong turns, discontinuities, and no faithful global plan. Case V reveals difficulty grounding abstract movement rules, producing inconsistent arrow trajectories. Case VI produces visually plausible motions along individual paths but fails to preserve or present the comparative information required for contrastive reasoning. Taken together, these examples indicate that the model can simulate locally coherent short traces but systematically fails at long-horizon planning, rule-grounded execution, and object-persistent manipulations. 

\begin{takeawaybox}
Veo-3 can produce locally coherent, short-horizon trace animations in simple, low-branching scenarios, but it does not reliably execute long-horizon plans or rule-grounded sequences. 
\end{takeawaybox}

\begin{figure}
    \centering
    \includegraphics[width=\linewidth]{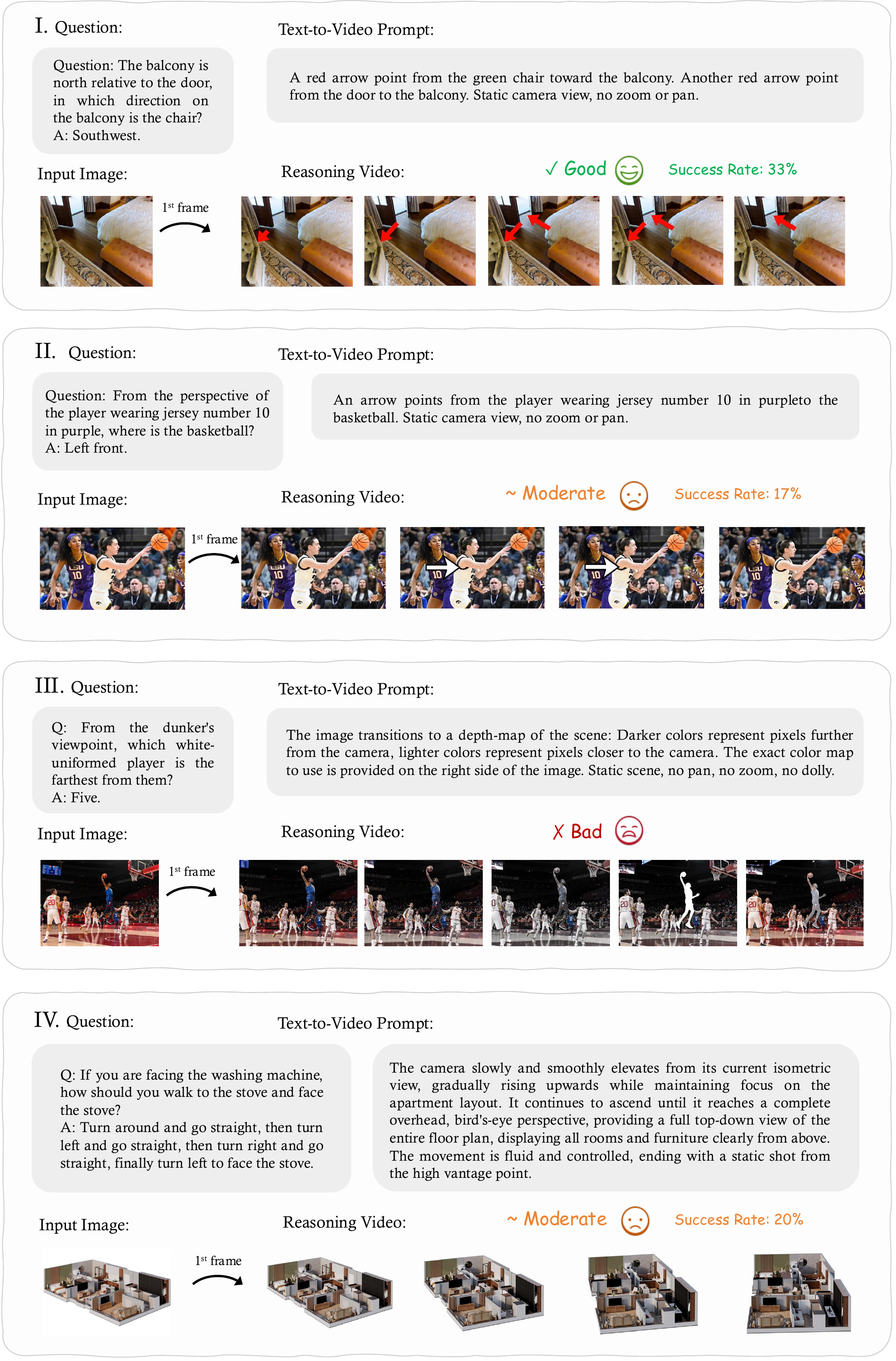}
    \caption{\textbf{Showcase of Real-World Spatial Reasoning by Veo-3.} Although Veo-3 can reason about simple spatial layouts, it still struggles to maintain consistency under complex perspective or orientation changes.}
    \label{fig:spatial}
\end{figure}

\subsection{Real-World Spatial Reasoning}
\paragraph{Task Description and Evaluated Aspects.}
This task investigates Veo-3~\cite{GoogleDeepMind2025Veo3}'s ability to perceive and maintain spatial relations within natural scenes, with a focus on reasoning about viewpoint change, orientation consistency, and reference-frame alignment. We assess whether the model preserves a stable global coordinate frame and coherent scene orientation under varying viewpoints, and whether objects retain correct relative positions and orientations with respect to each other across different views.

\paragraph{Definition of \good / \moderate / \bad.}
We define the evaluation criteria in three levels:
\begin{center}
\begin{itemize}[
    label={},
    leftmargin=1.5em,
    rightmargin=1.5em,
    itemsep=3pt, topsep=2pt
]
    \item \chgood: Scene orientation, reference frame, and viewpoint are consistent and correctly represent spatial relations. The camera remains steady and the motion is natural.
    \item \chmoderate: Scene roughly matches the instruction but contains small perspective errors, unnatural transitions, or partial mirroring. Motion remains interpretable but not physically coherent.
    \item \chbad: Reference frame or direction is wrong; viewpoint shifts abruptly or inconsistently. Video suffers from strong camera drift, disorienting motion, or spatial chaos.
\end{itemize}
\end{center}

\paragraph{Data Source.}
To evaluate on orientation and layout reasoning, we specifically sample data from \textit{MMSI-Bench}~\cite{yang2025mmsi}. Also, the tasks of perspective taking and spatial interaction are selected from the \textit{OmniSpatial} dataset~\cite{jia2025omnispatial}. 

\paragraph{Example and Analysis.}
As shown in~\Cref{fig:spatial}, Veo-3 can correctly handle basic spatial layouts in case I, but struggles with complex viewpoints or orientation changes in case II. The perspective transformations are sometimes inaccurate or even incorrect, suggesting that the model tends to prioritize visual plausibility over precise spatial reasoning, which hinders further reasoning in case IV. Moreover, case III demonstrates that Veo-3 has difficulty understanding depth, further limiting its spatial reasoning capability. 

\begin{takeawaybox}
While Veo-3 exhibits an emerging ability for simple real-world spatial reasoning, its capability remains insufficient for handling more complex spatial understanding tasks.
\end{takeawaybox}

\begin{figure}
    \centering
    \includegraphics[width=\linewidth]{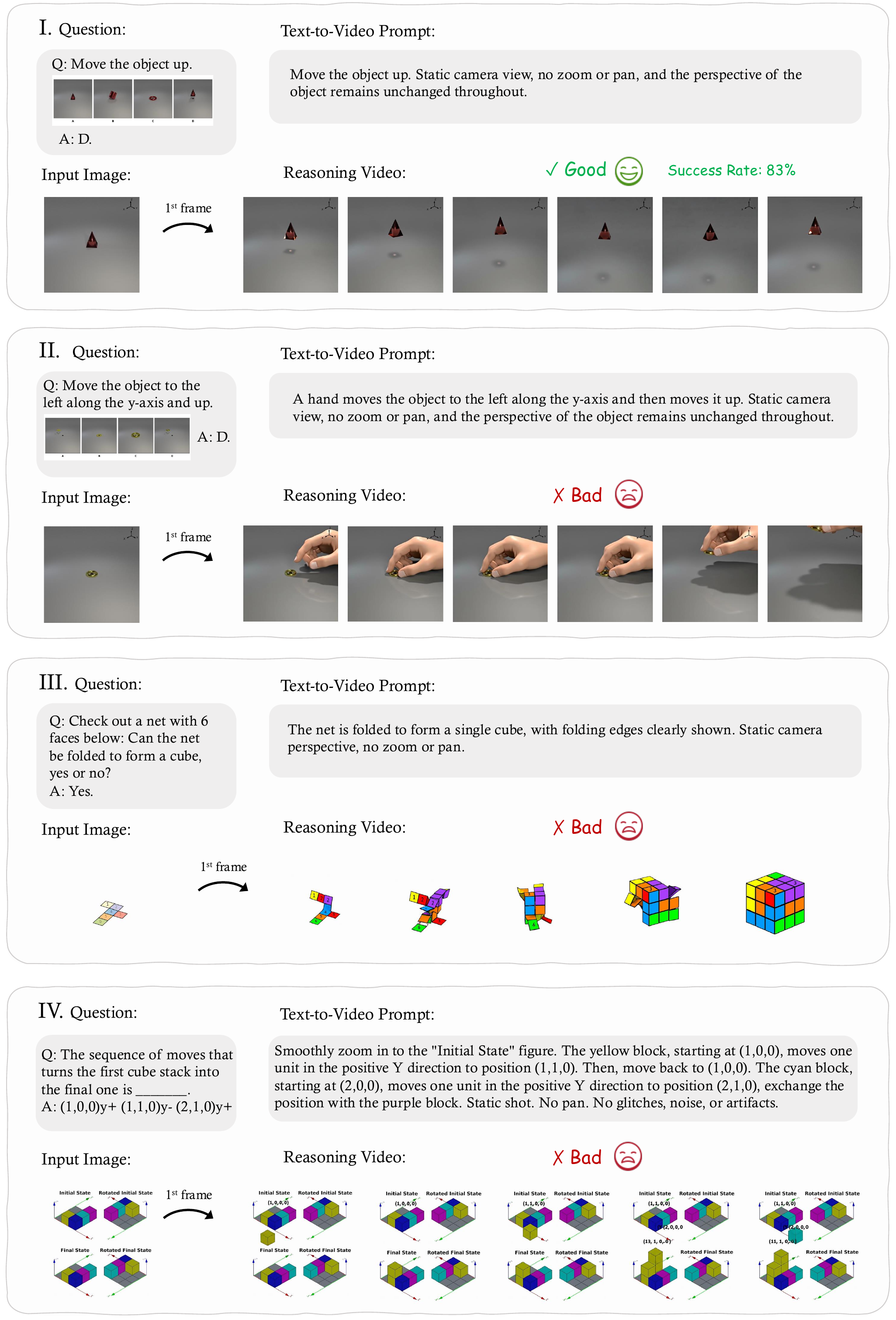}
        \caption{\textbf{Showcase of 3D Geometry Reasoning by Veo-3 (Part I).} While Veo-3 shows certain potential in basic 3D geometry reasoning, its performance remains unstable for complex geometry transformations.}
    \label{fig:3d_geo1}
\end{figure}

\begin{figure}
    \centering
    \includegraphics[width=\linewidth]{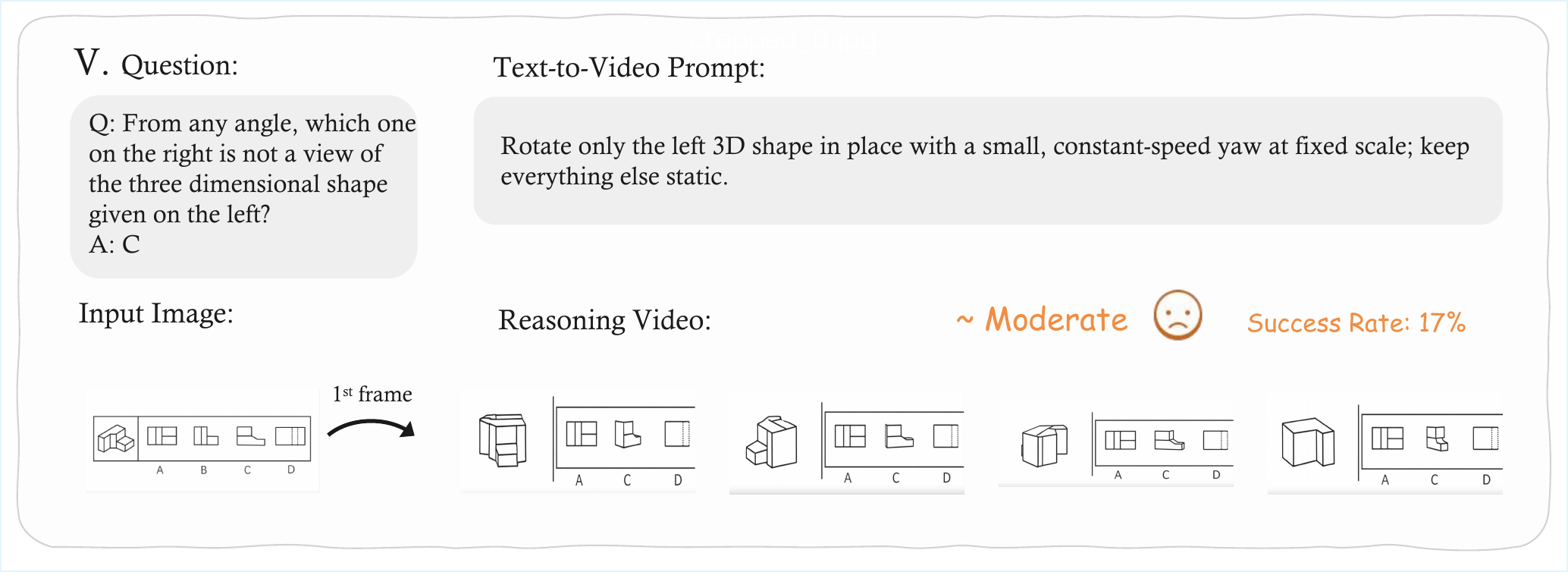}
    \caption{\textbf{Showcase of 3D Geometry Reasoning by Veo-3 (Part II).} The model often generates misaligned or self-intersecting structures, compromising geometric consistency.}
    \label{fig:3d_geo2}
\end{figure}

\subsection{3D Geometry Reasoning}

\paragraph{Task Description and Evaluated Aspects.}
We also evaluate Veo-3's potential on 3D geometry reasoning tasks, such as geometric object motion and three-dimensional structural transformations like reconstructing a cube net. The assessment focuses on three key dimensions: geometric accuracy, structural completeness throughout the transformation, and visual continuity across frames.

\paragraph{Definition of \good / \moderate / \bad.}
We categorize the model's performance into three levels:
\begin{center}
\begin{itemize}[
    label={},
    leftmargin=1.5em,
    rightmargin=1.5em,
    itemsep=3pt, topsep=2pt
]
    \item \chgood: Transformations like folding, rotation and assembly are geometrically correct, visually smooth, and continuous, maintaining structural integrity and realistic motion. No broken edges, jumps, or spatial artifacts.
    \item \chmoderate: Transformations are partially correct but show local misalignment, unrealistic deformation, or discontinuous motion; geometry is roughly interpretable but imperfect.
    \item \chbad: Transformation fails. For example, wrong fold, structure collapse, or impossible geometry. Motion is erratic, discontinuous, or visually implausible, breaking the sense of physical realism.
\end{itemize}
\end{center}

\paragraph{Data Source.}
To construct diverse and representative evaluation data, we adapt tasks from established geometric spatial reasoning datasets, including the \textit{3D-Text-Instruct} and \textit{Folding Nets} subsets of the \textit{STARE} benchmark~\cite{li2025unfolding}, the \textit{BlockMoving} subset from the \textit{SpatialViz-Bench}~\cite{wang2025spatialviz}, as well as \textit{VisuLogic}~\cite{xu2025visulogic} benchmark. 

\paragraph{Example and Analysis.}
We showcase the results of Veo-3 on 3D geometry reasoning tasks in~\Cref{fig:3d_geo1} and ~\Cref{fig:3d_geo2}. Veo-3 demonstrates a degree of potential on 3D geometry reasoning, performing reasonably well on simple, single-step geometric transformations, as shown in case I. However, its performance degrades noticeably when facing multi-step or compositionally complex transformations in case II. As presented in cases III and V, the model frequently produces misaligned or self-intersecting structures, leading to a loss of geometric consistency.
Further observations in case IV, show that while the model can partially understand the geometric shape of individual objects, it lacks a coherent understanding of coordinate systems and the spatial relationships among multiple objects.

\begin{takeawaybox}
Veo-3 exhibits emerging reasoning potential on basic 3D transformations but breaks down on complex or multi-step geometry, often yielding misaligned or self-intersecting structures. Its 3D geometric reasoning remains fragile, revealing substantial gaps in its ability to function as a reliable 3D geometry reasoner.
\end{takeawaybox}

\begin{figure}
    \centering
    \includegraphics[width=\linewidth]{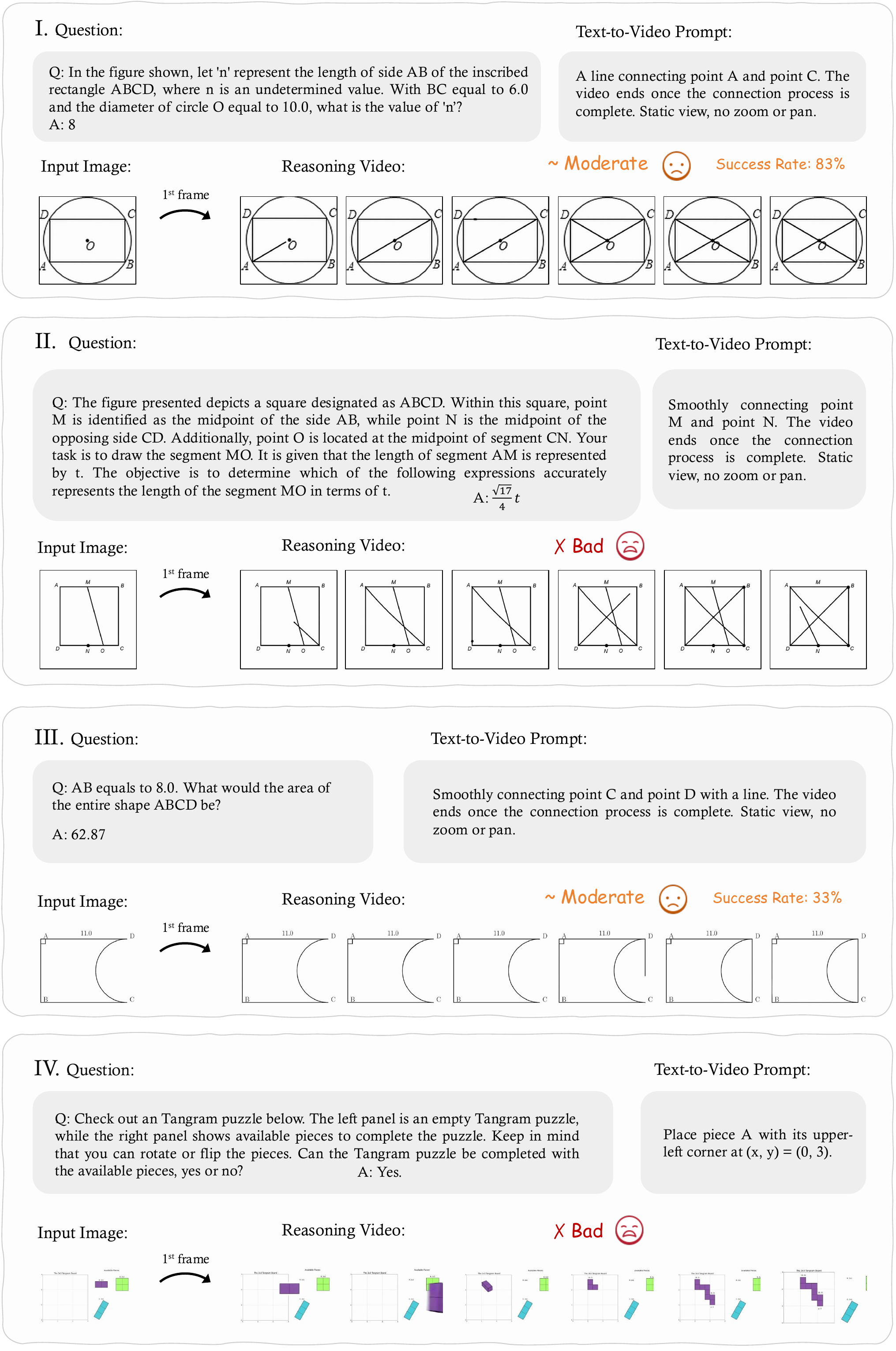}
    \caption{\textbf{Showcase of 2D Geometry Reasoning by Veo-3 (Part I).} While Veo-3 shows potential in recognizing simple patterns, it lacks the robust constraint awareness essential for accurate geometric manipulation.}
    \label{fig:2d_geo1}
\end{figure}

\begin{figure}
    \centering
    \includegraphics[width=\linewidth]{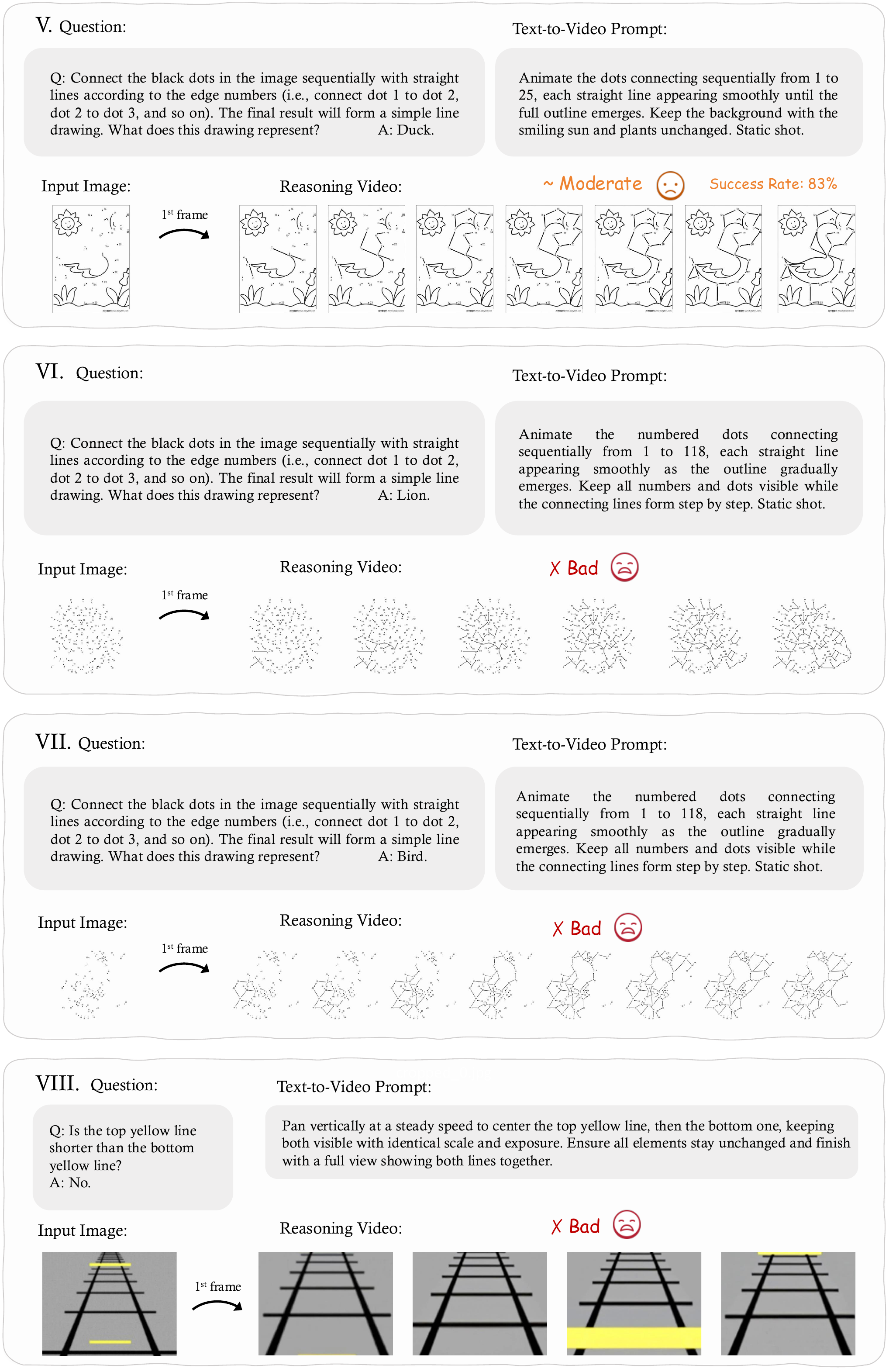}
    \caption{\textbf{Showcase of 2D Geometry Reasoning by Veo-3 (Part II).} Veo-3's reasoning abilities are further challenged by complex sequential instructions and the need to preserve structural integrity.}
    \label{fig:2d_geo2}
\end{figure}
\subsection{2D Geometry Reasoning}
\paragraph{Task Description and Evaluated Aspects.}
To assess a model's competence in 2D geometric reasoning, we evaluate its zero-shot performance on planar geometric construction tasks. These tasks involve drawing geometric relations by connecting points, adding auxiliary lines, and moving geometric shapes. The evaluation focuses on whether the generated constructions or movements accurately reflect the described geometric relationships and adhere to the given instructions, while maintaining smooth, stable operations that ensure visual clarity and coherence throughout the process.

\paragraph{Definition of \good / \moderate / \bad.}
We rate the performance according to the following criteria:
\begin{center}
\begin{itemize}[
    label={},
    leftmargin=1.5em,
    rightmargin=1.5em,
    itemsep=3pt, topsep=2pt
]
    \item \chgood: Constructions and movements are geometrically accurate and visually smooth. Endpoints, intersections, angles, and motion trajectories align correctly with the instructions. Both drawing and movement processes are stable, fluid, and natural, resembling human sketching or manipulation.
    \item \chmoderate: Constructions and movements roughly follow the intended geometry but exhibit minor inaccuracies in line placement, shape alignment, trajectory, or smoothness. Some local jitter or abrupt motion may appear, but the overall structure and motion remain interpretable.
    \item \chbad: Constructions or movements deviate substantially from geometric correctness. Lines or shapes may be misplaced, disconnected, or moved in a chaotic or discontinuous manner (\eg jittering, overlapping, or distorted paths), leading to visual instability and loss of interpretability.
\end{itemize}
\end{center}

\paragraph{Data Source.}
The evaluation data are drawn from multiple established sources, including the \textit{Geo170k} dataset~\cite{gao2023g}, the \textit{VarsityTutors} subset of \textit{Math-PUMA}~\cite{zhuang2025math} dataset, the \textit{line-connection} subset of \textit{RBench-V}~\cite{guo2025rbench}, the \textit{MAVIS-Gen}~\cite{zhang2024mavis}, \textit{Tangram Puzzle} subsets of the \textit{STARE}~\cite{li2025unfolding} benchmark, and data from \textit{VAT}~\cite{liu2025visual}.

\paragraph{Example and Analysis.}
The representative examples of the 2D geometry reasoning task are presented in~\Cref{fig:2d_geo1,fig:2d_geo2}. Veo-3 demonstrates a foundational capability for simple geometric connection tasks, correctly identifying and linking elements in straightforward scenarios like in case III. However, this basic competence is inconsistent. The model often prioritizes producing visually symmetric or semantically meaningful patterns rather than strictly adhering to geometric instructions (cases I and II). Furthermore, case II reveals instances where the model unintentionally modifies the original figures, indicating a limited awareness of geometric constraints and poor spatial consistency.
When tackling more complex connection tasks, the model frequently fails to interpret the intended drawing order or point indices, resulting in incorrect connection sequences, as demonstrated in cases V, VI, and VII. This is often coupled with an inability to control task termination, as the model tends to continue drawing beyond the required constructions. Finally, for tasks involving the movement of geometric shapes in cases IV and VIII, the model struggles to maintain geometric structural consistency throughout the motion.
\begin{takeawaybox}
Veo-3 shows initial 2D geometric reasoning ability but still falls short of consistent, constraint-aware geometric understanding, remaining far from a robust geometric reasoner.
\end{takeawaybox}

\begin{figure}
    \centering
    \includegraphics[width=\linewidth]{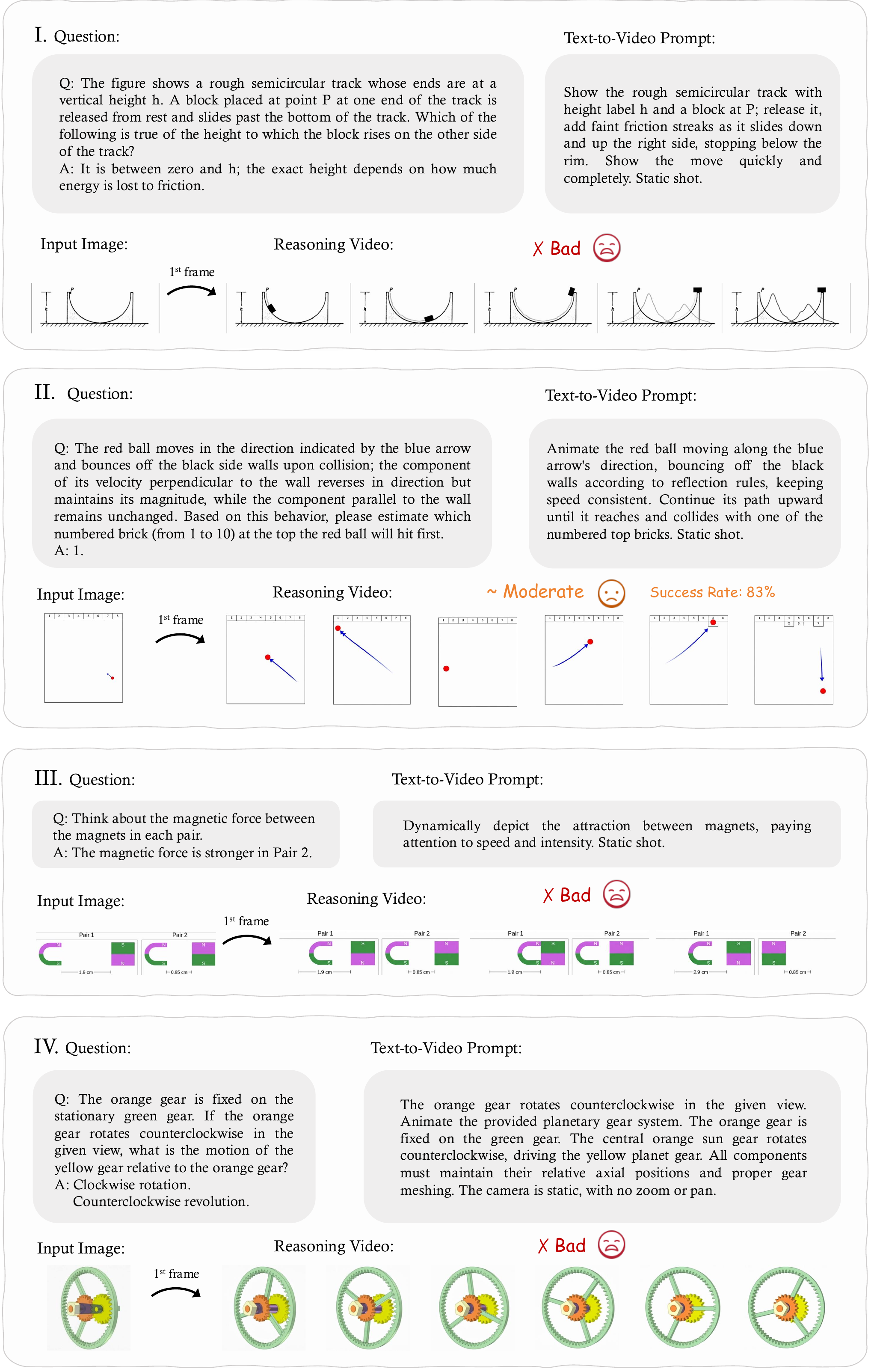}
    \caption{\textbf{Showcase of Physics-based Reasoning by Veo-3.} The physics scenarios demonstrate locally plausible dynamics and reflections, alongside systematic quantitative and causal inconsistencies under frictional, force-driven, or constrained interactions.}
    \label{fig:physics}
\end{figure}

\subsection{Physics-based Reasoning}
\paragraph{Task Description and Evaluated Aspects.}
The physics-based reasoning category assesses a model's capacity to depict and reason about motion dynamics, physical causality, and rule-based interactions between objects. 
Tasks in this group involve gravity, collisions, reflection, momentum, or energy conservation, requiring the model to generate physically plausible and temporally coherent motion. 
Evaluation focuses on two complementary aspects: (\textit{i})~physical plausibility, which means whether the simulated motion obeys common physical principles; and (\textit{ii})~causal correctness, which is whether object interactions are consistent with the underlying cause-and-effect relationships described in the prompt.

\paragraph{Definition of \good / \moderate / \bad.}
We rate the performance according to the following criteria:
\begin{center}
\begin{itemize}[
    label={},
    leftmargin=1.5em,
    rightmargin=1.5em,
    itemsep=3pt, topsep=2pt
]
    \item \chgood: The motion sequence adheres to physical laws such as gravity, momentum, and energy conservation. Object interactions are realistic and temporally smooth, and the visual outcome remains coherent and credible throughout.
    \item \chmoderate: The physical relations are approximately correct but include minor inconsistencies, such as irregular acceleration, timing mismatch, or slight violation of conservation. The overall motion remains interpretable and visually plausible.
    \item \chbad: The motion is physically implausible or visually chaotic—objects float, stop abruptly, or behave contrary to basic causal principles. Severe artifacts or temporal discontinuities disrupt the perception of a coherent physical process.
\end{itemize}
\end{center}

\paragraph{Data Source.}
We draw samples from \textit{MMMU}~\cite{yue2024mmmu}, \textit{ScienceQA}~\cite{lu2022learn}, and related physical reasoning subsets of \textit{RBench-V}~\cite{guo2025rbenchv} and \textit{SpatialViz-Bench}~\cite{wang2025spatialviz}, covering scenarios such as object collisions, pendulum motion, frictional sliding, and optical or magnetic interactions. 

\paragraph{Example and Analysis.} 
~\Cref{fig:physics} presents four representative physics tasks and their outputs. Case I shows that the model can produce a visually coherent slide, but the behavior violates basic physical laws. Case II is the most reliable, where reflections and general trajectory shape are rendered plausibly and the task attains a high success rate, although small angular or timing offsets are common. In case III, the model conveys attraction through motion, yet the depicted dynamics do not reliably track the intended force magnitudes or causal ordering. Finally, case IV exposes structural failures, incorrect meshing, inconsistent relative rotations, and nonphysical contact behavior occur frequently, so the mechanical constraints are not respected. Overall, the model can synthesize locally plausible dynamics and handle simple reflection rules, but it fails to maintain quantitative physical constraints and causal fidelity in frictional, force-driven, or mechanically constrained scenarios.

\begin{takeawaybox}
Veo-3 often generates visually plausible short-term dynamics, but it systematically fails to preserve quantitative physical constraints (energy, momentum), causal ordering, and contact mechanics in frictional, force-driven, or mechanically constrained scenarios. Thus, its outputs are somewhat useful for qualitative illustration but are not reliable for quantitative physics inference or causal prediction.
\end{takeawaybox}

\subsection{Rotation Reasoning}
\begin{figure}
    \centering
    \includegraphics[width=\linewidth]{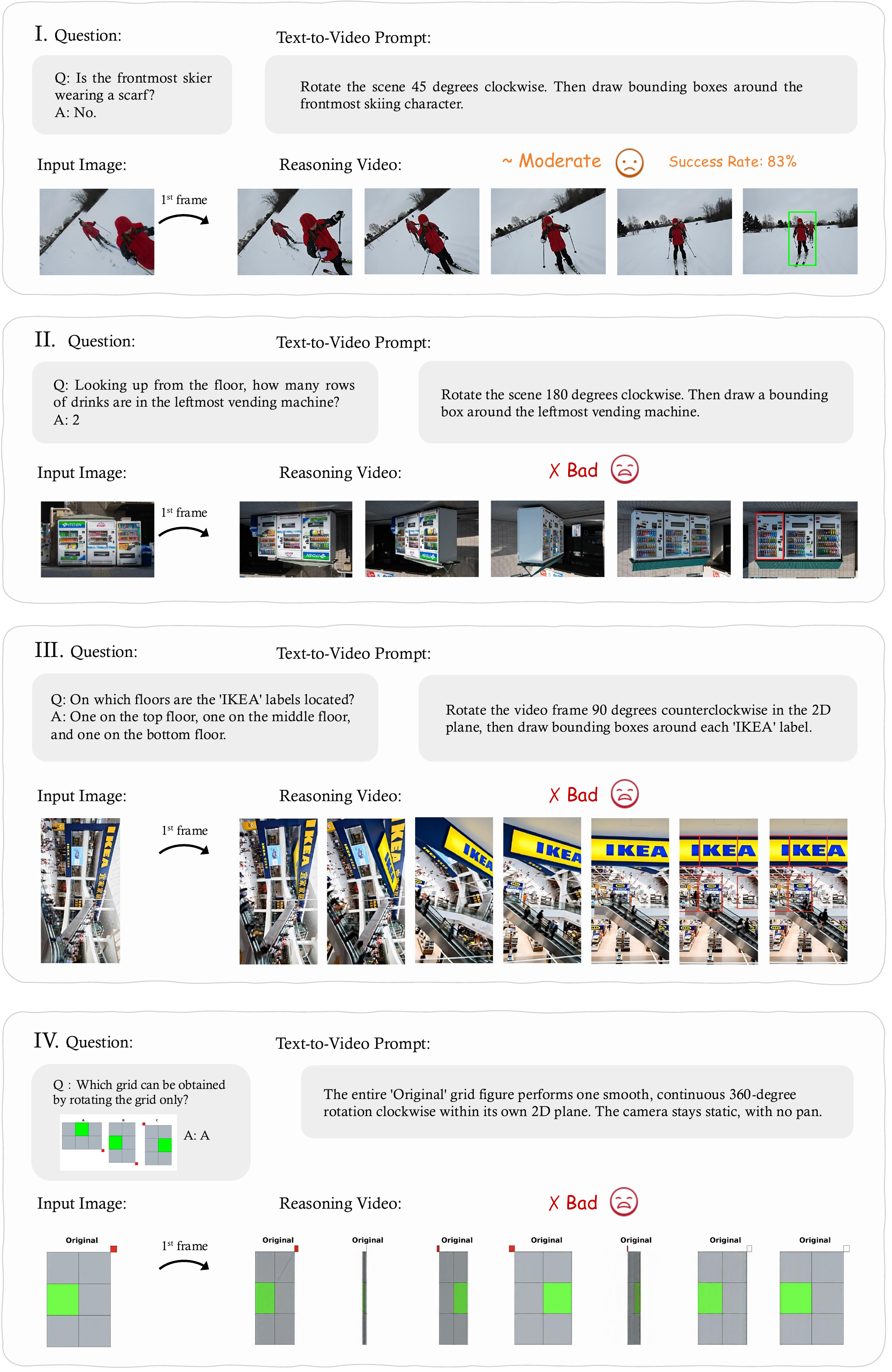}
    \caption{\textbf{Showcase of Rotation Reasoning by Veo-3.} Veo-3 struggles in complex scenes. However, its foundational grasp of simple rotations signals its potential to support rotation-based reasoning tasks.}
    \label{fig:rotation}
\end{figure}
\paragraph{Task Description and Evaluated Aspects.}
The rotation reasoning task assesses the ability to reason about planar object rotation and maintain consistent spatial grounding under rotational transformations, thereby supporting subsequent reasoning processes. In each instance, the model is required to accurately rotate target objects within a fixed 2D plane while preserving the overall scene structure and structural consistency, followed by performing reasoning tasks like grounding and OCR. The evaluation focuses on both the accuracy of the rotation in terms of angle and direction, and the precision of the resulting reasoning tasks.

\paragraph{Definition of \good / \moderate / \bad.}
Model outputs are categorized into three quality levels:
\begin{center}
\begin{itemize}[
    label={},
    leftmargin=1.5em,
    rightmargin=1.5em,
    itemsep=3pt, topsep=2pt
]
    \item \chgood: The rotation is accurate, complete, and strictly confined to the 2D plane, with no extraneous scene motion. The following reasoning tasks are completed correctly. Target objects remain precisely grounded after rotation.
    \item \chmoderate: The rotation is largely correct but may be incomplete or slightly off-angle, though still confined to the 2D plane. The following reasoning tasks are mostly completed. Minor temporal or visual inconsistencies may appear, but do not alter the core 2D structure or object grounding.
    \item \chbad: The model fails to perform the correct rotation, extends the transformation into 3D space, or introduces substantial scene distortion. Cannot complete the following reasoning task. The original 2D structure is altered, leading to inaccurate grounding of the target objects.
\end{itemize}
\end{center}

\paragraph{Data Source.}
To specifically assess the rotation reasoning task, we recruit some PhD-level experts with deep expertise in text-image reasoning to design the evaluation data manually, followed by the necessary review process, as mentioned in~\Cref{sec:review}. Each question is designed following the principle that it must involve a 2D rotation to reach the correct solution, ensuring the task genuinely probes rotational understanding rather than simple visual matching. Moreover, we sample data from the \textit{2DRotation} subset from the \textit{SpatialViz-Bench}~\cite{wang2025spatialviz}, and reformulate the question into instructions for the video models.

\paragraph{Example and Analysis.}
The results are shown in~\Cref{fig:rotation}. In case I, we find that Veo-3 handles small-angle rotations and simple planar scenes reasonably well, demonstrating a basic grasp of rotational motion. However, in more complex scenarios like cases II, III, and IV, the model often ignores the 2D rotation constraint and inadvertently alters the 3D structure, resulting in incorrect rotations and degraded spatial grounding. Such errors frequently propagate to downstream tasks, such as OCR in case III, or object localization in case II, due to inconsistencies in post-rotation alignment. These observations suggest that the reasoning behavior of Veo-3 remains more pattern-driven rather than principle-driven. However, as it demonstrates a partial understanding of planar rotation, this can to some extent facilitate subsequent reasoning tasks.
\begin{takeawaybox}
Veo-3 exhibits only a superficial understanding of rotation reasoning. While it can approximate small planar rotations, it fails to preserve geometric consistency under larger or compound transformations.
\end{takeawaybox}

\subsection{Table and Chart Reasoning}
\paragraph{Task Description and Evaluated Aspects.}
The table and chart reasoning task requires the model to identify and focus on the key elements within visualizations or tabular data. For evaluation, we further consider how effectively the model identifies the regions relevant to the query and whether it can transition smoothly and visually coherently to these areas, preserving clarity, continuity, and proper scaling.

\paragraph{Definition of \good / \moderate / \bad.} 
We rate the performance according to the following criteria:
\begin{center}
\begin{itemize}[
    label={},
    leftmargin=1.5em,
    rightmargin=1.5em,
    itemsep=3pt, topsep=2pt
]
    \item \chgood: Camera precisely focuses on the correct chart or table segment, smoothly highlighting or zooming into the queried data (\eg correct year, category, or value). Motion is continuous, the chart and table remain clear, and no distortion or overexposure occurs.
    \item \chmoderate: Camera approximately focuses on the right region but partially misses boundaries, introduces slight blur, or transitions abruptly. Data can still be inferred.
    \item \chbad: Video fails to locate the correct region or changes the chart or table geometry unnaturally. Motion jitter, scaling errors, or artifacts make data unreadable or misleading.
\end{itemize}
\end{center}

\begin{figure}
    \centering
    \includegraphics[width=\linewidth]{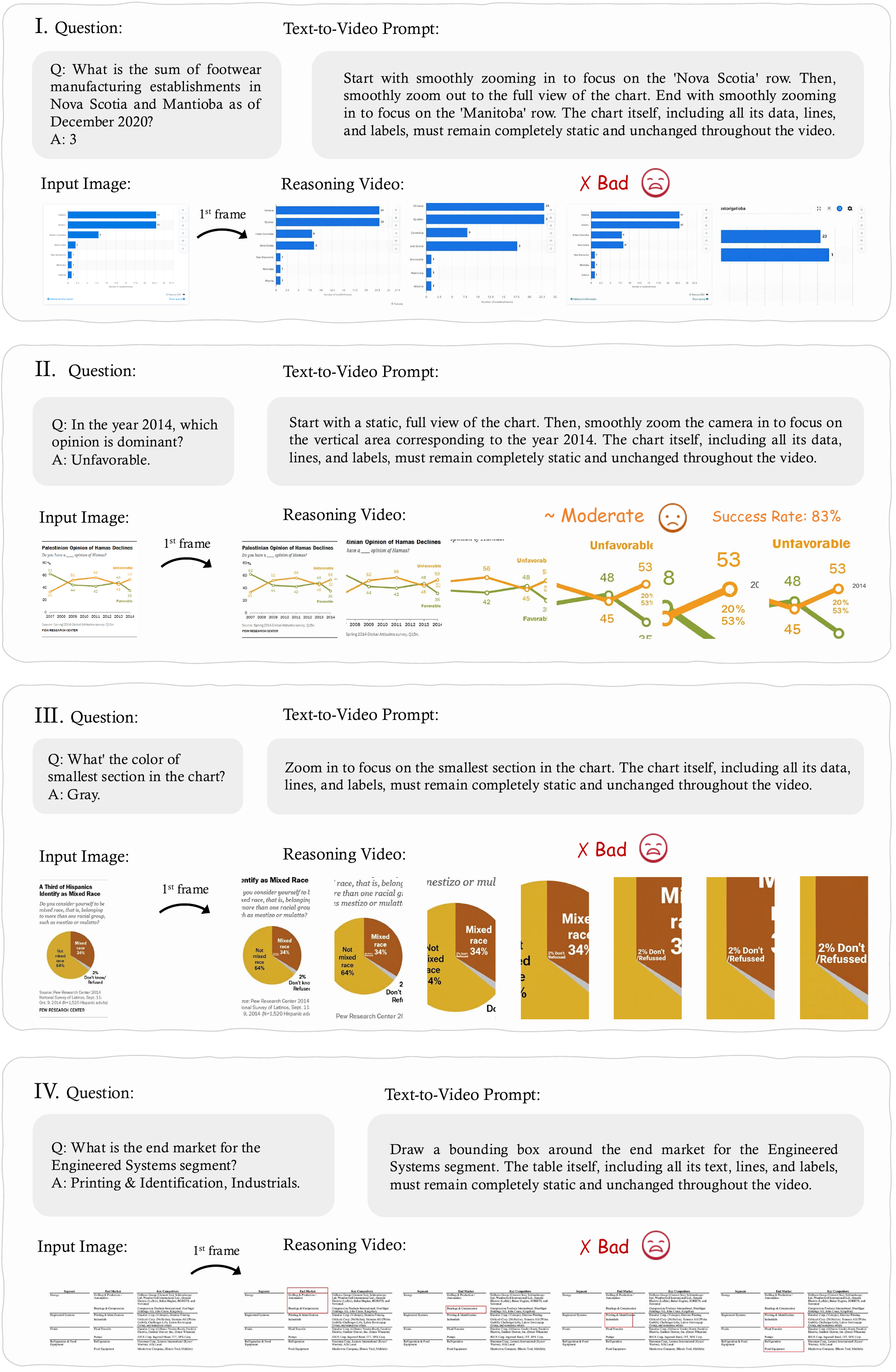}
    \caption{\textbf{Showcase of Table and Chart Reasoning by Veo-3.} Veo-3 demonstrates an initial ability to focus on relevant data regions but lacks the precision and consistency required for reliable visual analysis.}
    \label{fig:chart_table}
\end{figure}

\paragraph{Data Source.}
We use samples from the \textit{ChartQA}~\cite{masry-etal-2022-chartqa} dataset and \textit{TableVQA-Bench}~\cite{kim2024tablevqa}. 

\paragraph{Example and Analysis.}
For charts, as presented in cases I, II and III in~\Cref{fig:chart_table}, Veo-3 can often zoom into an approximately correct region but lacks the precision needed to accurately locate the queried data. For tables, as shown in case IV, Veo-3 fails to correctly identify the required element and tends to select entries randomly. The model also frequently adds, modifies, or distorts existing chart and table elements, resulting in visual inconsistencies that undermine the accuracy of chart interpretation.
\begin{takeawaybox}
Veo-3 demonstrates emerging competence and potential in structured visual understanding, but still falls short of functioning as a precise and reliable chart-table reasoner.
\end{takeawaybox}

\subsection{Object Counting Reasoning}
\paragraph{Task Description and Evaluated Aspects.}
In this category, we focus on the ability to accurately enumerate objects within a 2D or 3D scene. In each instance, the model is required to identify, ground, and count target objects, typically by highlighting, drawing bounding boxes, applying numerical labels, or panning. The evaluation focuses on the accuracy of the count and the precision of the spatial grounding, performed within a scene that remains static or experiences only minimal motion, ensuring the counting process is not influenced.

\paragraph{Definition of \good / \moderate / \bad.}
Model outputs are categorized into three quality levels:
\begin{center}
\begin{itemize}[
    label={},
    leftmargin=1.5em,
    rightmargin=1.5em,
    itemsep=3pt, topsep=2pt
]
    \item \chgood: The model precisely highlights, draws bounding boxes around, or labels the objects with correct numbers, and performs smooth and controlled panning when necessary to cover all targets. Motion is continuous, and the scene remains static or experiences only slight changes that do not influence the counting process.
    \item \chmoderate: The model approximately highlights or draws bounding boxes around the objects, or performs panning with minor instability or incomplete coverage. Objects or the scene may move or change slightly, but this does not strongly affect the counting process.
    \item \chbad: The model fails to correctly highlight, label, or draw bounding boxes around the objects, or pans erratically such that parts of the scene are missed or revisited unnecessarily. Objects or the scene move or change substantially, severely affecting the counting process.
\end{itemize}
\end{center}

\begin{figure}
    \centering
    \includegraphics[width=\linewidth]{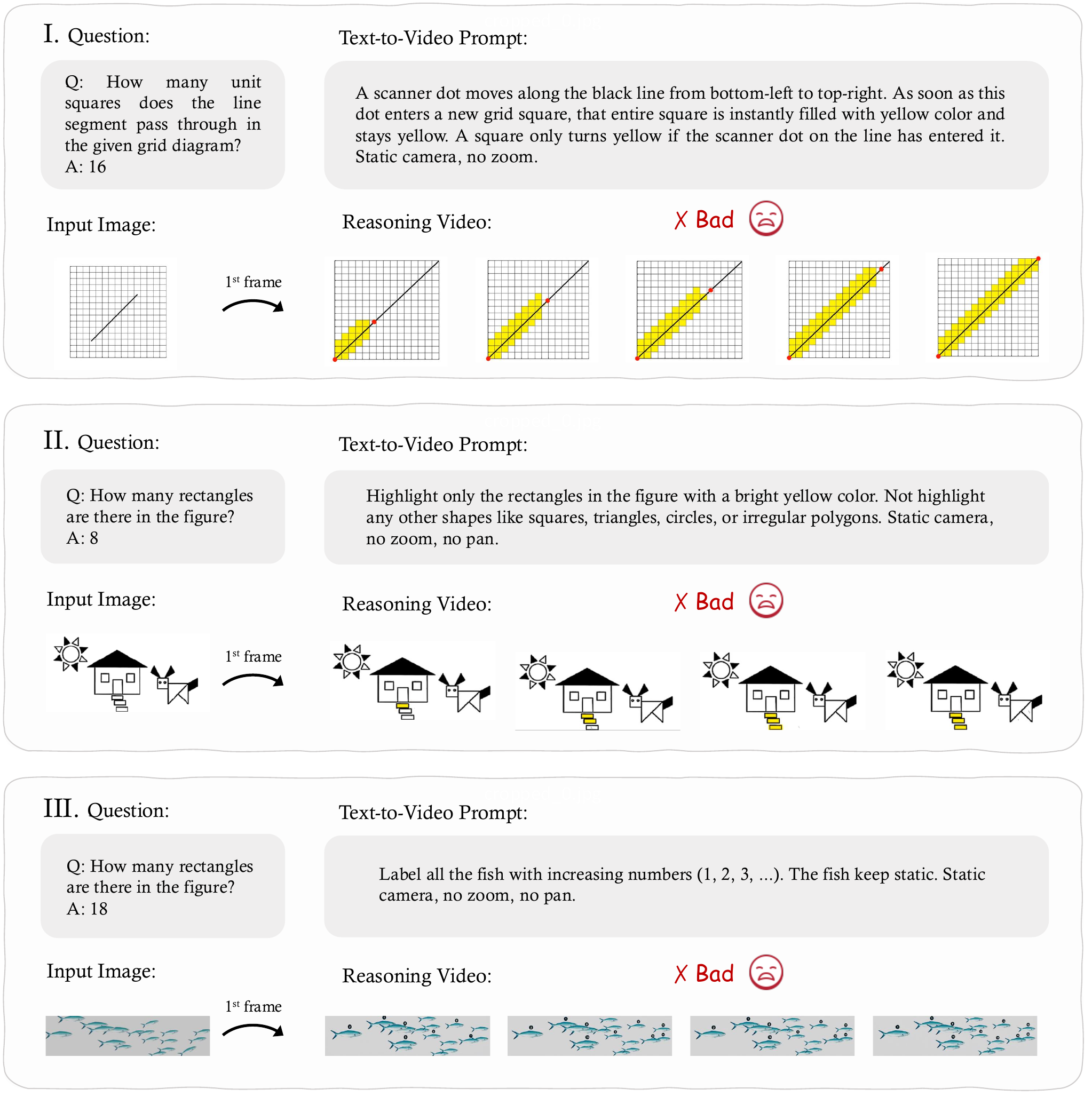}
    \caption{\textbf{Showcase of 2D Object Counting Reasoning by Veo-3.} Veo-3's lack of spatial control often introduces object motion, undermining the stability and accuracy of the counting process.}
    \label{fig:counting1}
\end{figure}

\begin{figure}
    \centering
    \includegraphics[width=\linewidth]{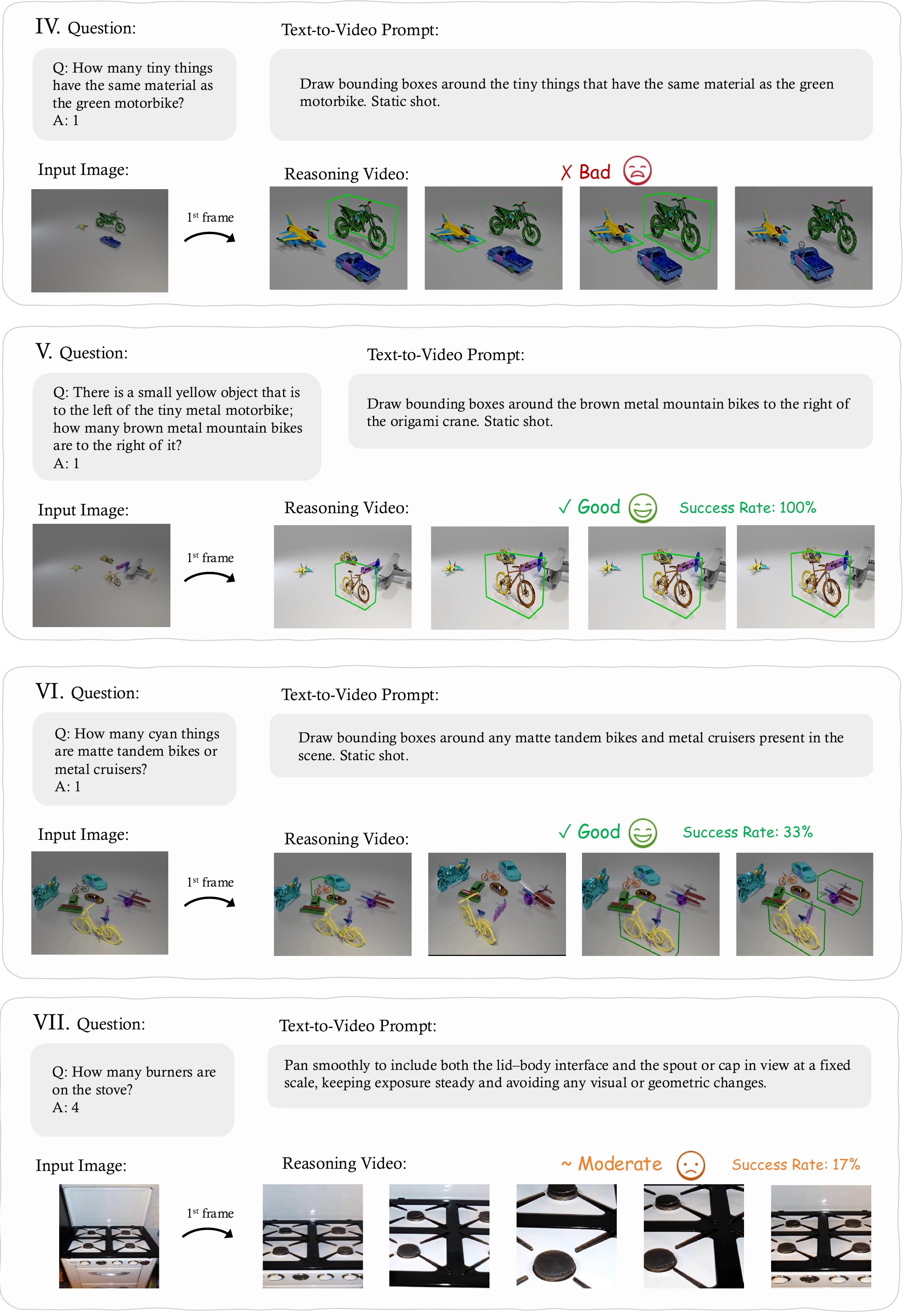}
    \caption{\textbf{Showcase of 3D Object Counting Reasoning by Veo-3.} Veo-3's basic 3D counting abilities are challenged by complex materials, geometric variations, and imprecise camera control.}
    \label{fig:counting2}
\end{figure}

\paragraph{Data Source.}
The 2D object counting data are sampled from the \textit{counting} subset of \textit{RBench-V}~\cite{guo2025rbench}. The 3D object counting data are from the \textit{Super-CLEVER} dataset~\cite{li2023super} and \textit{VAT}~\cite{liu2025visual}. 

\paragraph{Example and Analysis.}
The results are shown in~\Cref{fig:counting1,fig:counting2}. In the 2D counting tasks from cases I to III, objects frequently move or change during the process, negatively impacting counting stability and accuracy. In the 3D counting tasks, Veo-3 successfully handles simple grounding and counting scenarios, as demonstrated in case V, but struggles with scenes involving complex materials or geometric variations in cases VI and VII, leading to inaccurate counts. Additionally, in the panning process of case VII, the camera fails to precisely move to the regions containing all target objects, further hindering the counting process.

\begin{takeawaybox}
Veo-3 demonstrates basic counting capability but lacks the spatial control and robustness required for reliable object enumeration in dynamic or complex scenes.
\end{takeawaybox}
    
\subsection{GUI Reasoning}
\paragraph{Task Description and Evaluated Aspects.}
In the Graphical User Interface (GUI) reasoning task, we focus on the capability to understand and interact with graphical user interfaces across different operating systems, including Android, Linux, and Web environments. In each instance, the model is required to perform actions, such as clicking on specific UI elements. The evaluation focuses on the accuracy of the click and the temporal coherence of the interaction, ensuring the scene and irrelevant UI elements remain consistent.

\paragraph{Definition of \good / \moderate / \bad.}
We define the evaluation criteria in three levels:
\begin{center}
\begin{itemize}[
    label={},
    leftmargin=1.5em,
    rightmargin=1.5em,
    itemsep=3pt, topsep=2pt
]
    \item \chgood: The click is precise, with no extraneous actions. No superfluous icons appear, and the original data and icons remain unchanged.
    \item \chmoderate: The click is precise but may be accompanied by minor extraneous actions. Superfluous icons might appear but do not obscure the click target, and original data or icons show only slight alterations.
    \item \chbad: The click is imprecise or erratic. Original data and icons are significantly altered, hindering judgment and assessment.
\end{itemize}
\end{center}

\begin{figure}[t]
    \centering
    \includegraphics[width=\linewidth]{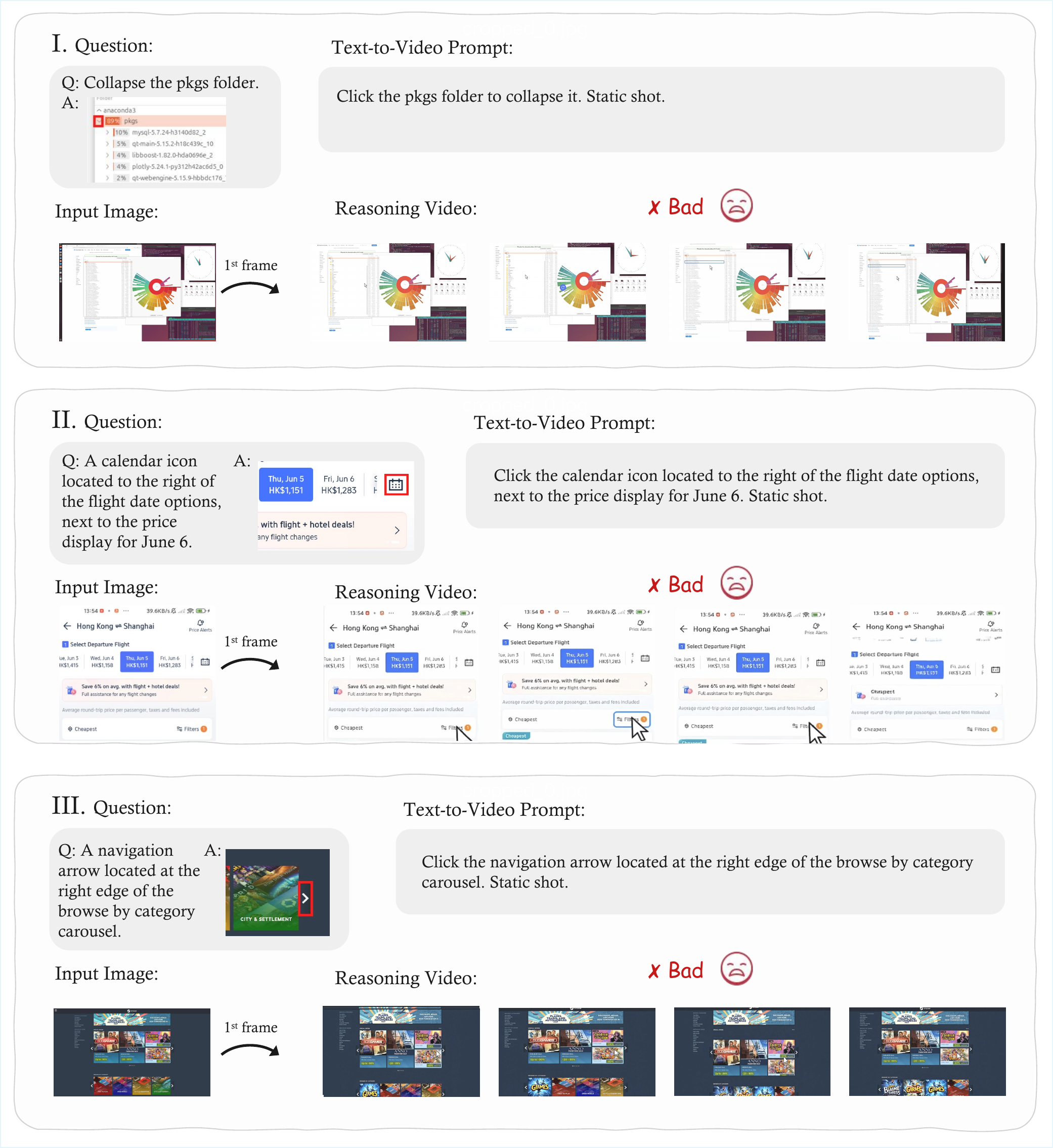}
    \caption{\textbf{Showcase of GUI Reasoning by Veo-3.} Veo-3’s attempts at graphical interface interaction exhibit visual inconsistencies and logical inaccuracies, indicating only a shallow grasp of underlying GUI logic. Note that the answer to each question is a bounding box. For visual clarity, screenshots with the ground-truth bounding boxes are shown.}
    \label{fig:gui}
\end{figure}

\paragraph{Data Source.}
The Linux data are selected from the \textit{Common Linux Screenshot} subset of \textit{ScreenSpot-Pro}~\cite{li2025screenspot}, while the Android and Web data are drawn from the \textit{OS Android} and \textit{OS Web} subsections of \textit{MMBench-GUI}~\cite{wang2025mmbench}, respectively.

\paragraph{Example and Analysis.}
Across the three cases in \Cref{fig:gui}, Veo-3 fails to accurately capture the correct click position and often exhibits inconsistencies between the click location and the resulting on-screen effect. In addition, it occasionally alters or generates new icons and text, which can interfere with judgment. In the Web system in case III, however, the model demonstrates partial GUI responsiveness and provides some degree of visual feedback.
\begin{takeawaybox}
Veo-3 demonstrates a limited awareness of GUI click actions, imitating interaction behaviors without fully grasping the underlying functional logic.
\end{takeawaybox}

\subsection{Embodied Reasoning}

\begin{figure}[t]
    \centering
    \includegraphics[width=\linewidth]{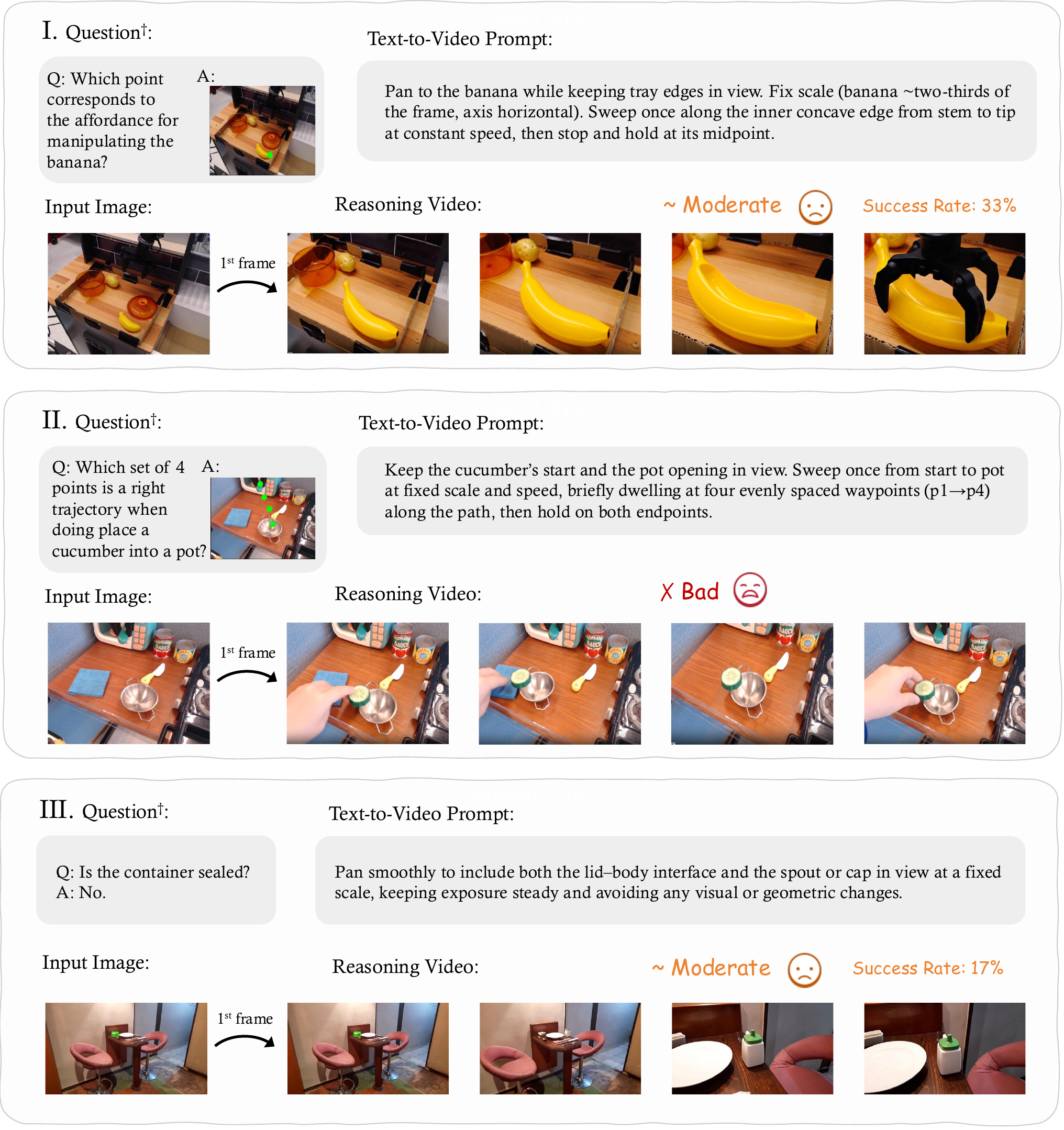}
    \caption{\textbf{Showcase of Embodied Reasoning by Veo-3.} It illustrates plausible static affordance detection in simple settings, common workaround/hallucination behaviors for dynamic manipulations, and failures to reliably localize or preserve manipulation-relevant context. $^\dagger$ Green points in the answer image denote ground-truth points or trajectories.}
    \label{fig:embodied}
\end{figure}

\paragraph{Task Description and Evaluated Aspects.} 
This category evaluates the model’s potential to perceive and reason about object affordances and manipulation dynamics. It involves recognizing both static and dynamic affordances, as well as identifying manipulation-relevant object and scene attributes. Evaluation focuses on two aspects: \textit{(i)} the generation of stable and contextually relevant visual sequences, and \textit{(ii)} the maintenance of reasoning fidelity without resorting to implausible planning shortcuts or hallucinated interactions.

\paragraph{Definition of \good / \moderate / \bad.}
We define the evaluation criteria in three levels:
\begin{center}
\begin{itemize}[
    label={},
    leftmargin=1.5em,
    rightmargin=1.5em,
    itemsep=3pt, topsep=2pt
]
    \item \chgood: The sweep/framing covers all candidates fairly (equal or near-equal dwell), centers the manipulation-relevant geometry (\eg handle + frame/gap, lid-body interface, hinge side) with crisp focus and stable scale; no cropping of key context; no content alterations. 
    \item \chmoderate: The view roughly includes the right region(s) but with minor bias or coverage issues: slight off-center, brief under-exposure of one candidate, small motion jitter, or shallow context (still enough to infer). 
    \item \chbad: The camera misses or biases the evidence (\eg lingers only on one point, crops away the hinge/rail, over-zooms a non-relevant patch), introduces distortion/content edits, or produces footage from which a fair decision cannot be made.
\end{itemize}
\end{center}

\paragraph{Data Source.} We select samples from \textit{Robobench}~\cite{luo2025robobenchcomprehensiveevaluationbenchmark} for the analysis. In addition to a general understanding of static attributes, we also sample data to assess whether Veo-3 can perform direct reasoning on tasks involving the generation of static and dynamic affordances. 
\paragraph{Example and Analysis.}
As shown in ~\Cref{fig:embodied}, Veo-3 demonstrates the ability to comprehend objects within real-world scenes. However, its capacity for assisting visual reasoning in embodied scenarios remains constrained by insufficient stability. As illustrated in case I, when provided with a clearly defined object for manipulation, Veo-3 is capable of generating plausible manipulation affordances. When it comes to dynamic affordances, Veo-3 tends to employ workarounds to compensate for its planning deficiencies, as evidenced in case II, where it generated a new cucumber instead of the intended object. With respect to static attributes, Veo-3 struggles to accurately differentiate visual prompts and misidentifies the position of containers. As shown in case III, the green box, intended to specify the location of the container, inadvertently led Veo-3 to produce hallucinations.

\begin{takeawaybox}
Veo-3's capabilities are currently limited to basic object recognition rather than true embodied reasoning. It lacks the necessary planning and stability to reliably interpret and act upon dynamic or spatially constrained instructions, indicating its limitations in understanding and reasoning of real-world interactions.
\end{takeawaybox}

\begin{figure}[t]
    \centering
    \includegraphics[width=\linewidth]{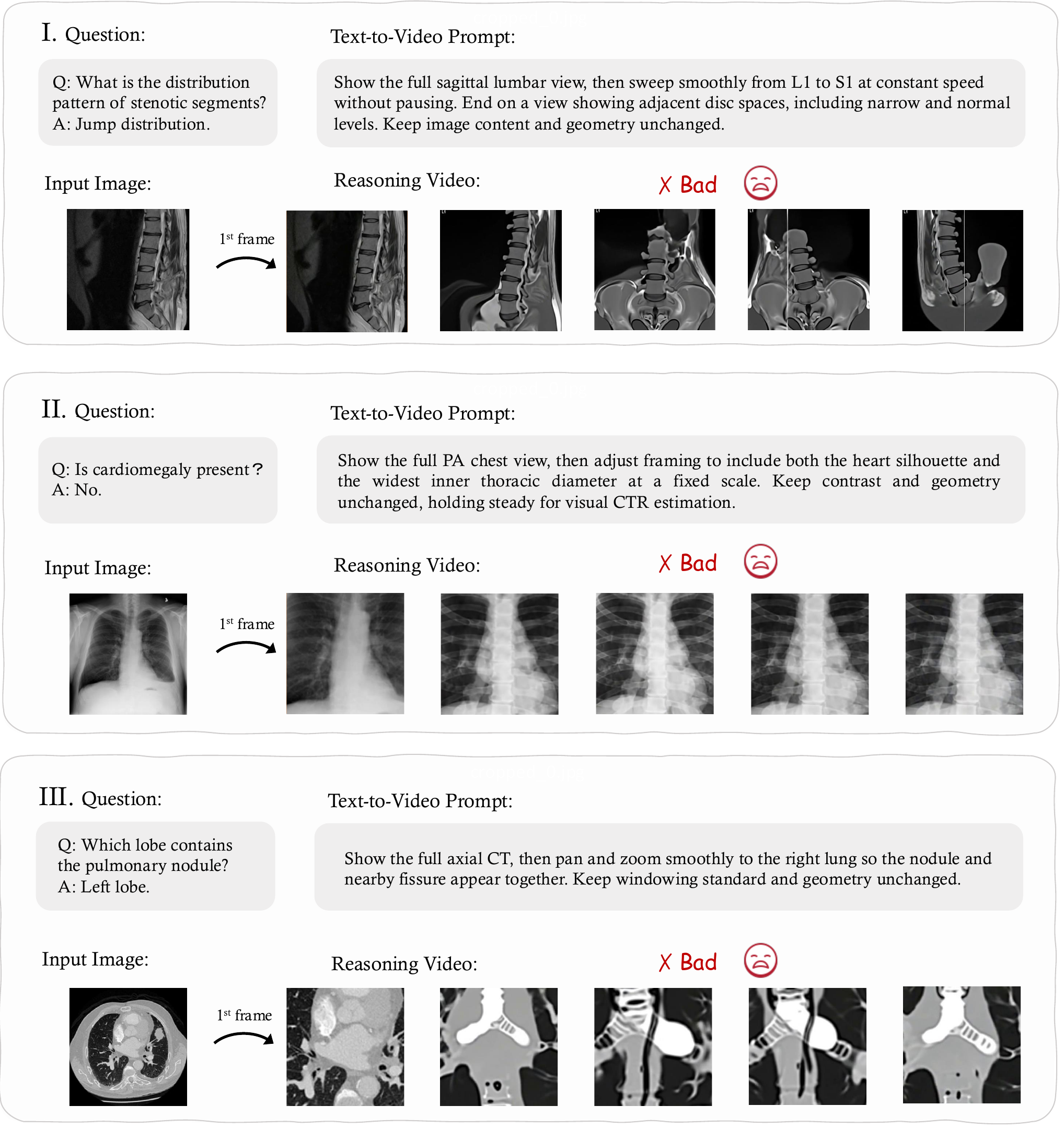}
    \caption{\textbf{Showcase of Medical Reasoning by Veo-3.} As shown in cases I and III, Veo-3 fails to maintain the shape of the rest of medical organization. Veo-3 also can not understand and precisely locate the mentioned medical terminology in the prompt, as demonstrated in case II.}
    \label{fig:medical}
\end{figure}

\subsection{Medical Reasoning}
\paragraph{Task Description and Evaluated Aspects.}
This category assesses the model’s ability to localize lesions or structures, identify relevant attributes (\eg side, lobe), recognize pathological patterns (\eg ``jump distribution”), and make binary decisions (\eg presence or absence). The evaluation focuses on both the correctness of object manipulation and the visual stability of the surrounding regions.

\paragraph{Definition of \good / \moderate / \bad.}
We define the evaluation criteria in three levels:
\begin{center}
\begin{itemize}[
    label={},
    leftmargin=1.5em,
    rightmargin=1.5em,
    itemsep=3pt, topsep=2pt
]
    \item \chgood: The camera cleanly settles on the correct anatomical level/lesion, with clear margins and readable context; motion is reasonable; no geometric distortion or content alteration. 
    \item \chmoderate: The view roughly covers the right area but is slightly off (partial coverage, mild blur, small framing mistakes). The general shape of the tissue or organ can still be observed. 
    \item \chbad: The video misses the target region or introduces distortions/crops that hide key cues. Tissues or organs begin to distort. Misleading results due to confusion of medical terminology.
\end{itemize}
\end{center}

\paragraph{Data Source.} We select samples representing different body parts from the \textit{ViTAR}~\cite{chen2025think} dataset.

\paragraph{Example and Analysis.} We showcase the evaluation results in ~\Cref{fig:medical}. Veo-3 retains the ability to manipulate images when dealing with medical images. However, due to its lack of medical knowledge, Veo-3 struggles to accurately manipulate the correct objects when instructions include medical terminology. This phenomenon is evident across all cases. Furthermore, Veo-3 cannot model medical organs effectively. When performing operations such as zooming in, the medical images suffer from significant distortion, resulting in a substantial loss of detail.

\begin{takeawaybox}
Veo-3's failure to handle the reasoning in the medical domain, causing distortion even on simple zoom-ins, highlights its limited grasp of specialized, non-general knowledge. 
\end{takeawaybox}

\section{\textsc{MME-CoF}}
\subsection{Benchmark Overview}
To standardize the empirical study and systematically evaluate the reasoning potential of \textit{state-of-the-art} generative video models~\cite{GoogleDeepMind2025Veo3,openai2024sora,openai2025sora2}, we introduce \textsc{MME-CoF}, which, to our knowledge, is the \textit{first} benchmark specifically designed to reveal and quantify the reasoning potential of video models.

\subsection{Benchmark Composition}

\begin{figure*}[t]
\vspace{-0.2cm}
\centering
\begin{minipage}{0.33\textwidth}
    \centering
    \includegraphics[width=\linewidth]{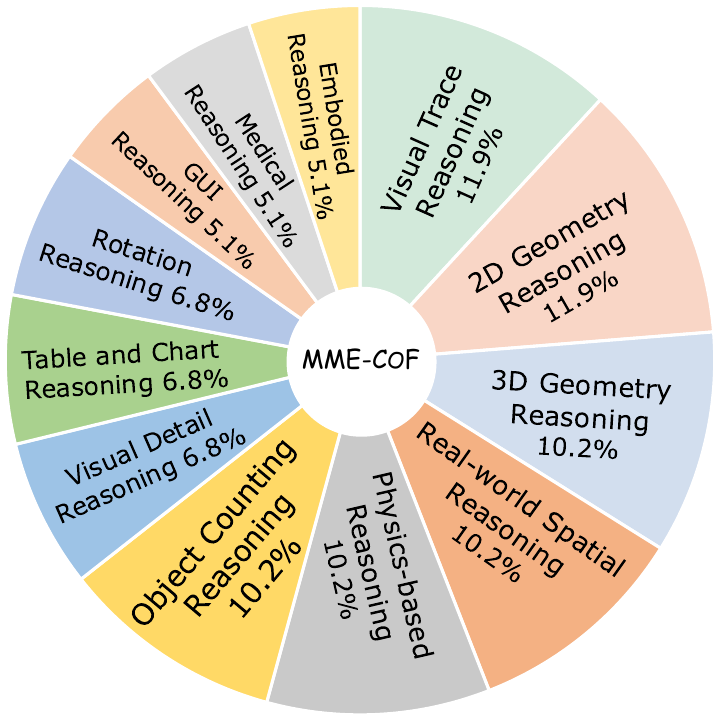}
    \vfill
    \caption{Category Distribution.}
    \label{fig:category}
\end{minipage}
\hspace{0.45cm}
\begin{minipage}{0.50\textwidth}
    \centering
    \small
    \captionof{table}{\textbf{Key Statistics of \textsc{MME-CoF}.}}
    \label{tab:dataset_stats}
    \vspace{0.2cm}
    \begin{tabular}{lc}
    \toprule
    Statistic & Number \\
    \midrule
    Total entries & 59 \\
    Total categories & 12 \\
    Max prompt length & 124 \\
    Avg prompt length & 36.7 \\
    Max entries per category & 7 \\
    Avg entries per category & 4.9 \\
    \bottomrule
    \end{tabular}
\end{minipage}
\label{fig:mme_cof_overview2}
\end{figure*}

\paragraph{Data Curation and Distribution.}
Aligning with the task taxonomy in \Cref{sec:analysis_overview}, the \textsc{MME-CoF} benchmark is curated from the cases used in our empirical study. 
It comprises 59 curated entries and instruction prompts spanning 12 diverse reasoning categories. The key statistics of \textsc{MME-CoF} and its overall composition are summarized in Table~\ref{tab:dataset_stats}, ~\Cref{fig:overview:wc} and ~\Cref{fig:category}.

\paragraph{Review Process.}
\label{sec:review}
Following the prompt design protocol in \Cref{sec:analysis_overview}, all prompts undergo a two-stage review process. In the \textit{cross-validation} phase, each prompt was independently reviewed by another expert to ensure semantic clarity, alignment with the intended reasoning task, and the absence of linguistic bias. In the \textit{final adjudication} phase, discrepancies were jointly discussed and resolved through consensus. 
This multi-step procedure ensured that every prompt was conceptually precise, visually grounded, and fully aligned with the evaluation objectives of \textsc{MME-CoF}.

\subsection{Evaluation Protocol}
\paragraph{Models and Generation Settings.} We evaluate the leading video models in a zero-shot setting, including Kling-v1~\cite{kuaishou2024kling}, Seedance-1.0-pro~\cite{gao2025seedance}, Veo-3.0-preview~\cite{wiedemer2025video}, Veo-3.0-fast~\cite{wiedemer2025video}, Sora-2~\cite{openai2025sora2}, Sora-2-pro~\cite{openai2025sora2}. Each model generates six video samples per prompt, and final scores were computed as the mean across all samples. All videos are generated at a 16:9 aspect ratio. We adopt the default 8-second duration for the Sora and Veo series, while retaining the default 5-second length for Kling and Seedance. 
Note that, since most video models apply automated safety filters and content moderation, which may block sensitive content, we exclude videos that are suppressed by such filters from our evaluation. 

\paragraph{Evaluation Metrics.} 
We employ Gemini-2.5-Pro~\cite{comanici2025gemini} as an automatic verifier to evaluate each generated video. Gemini is prompted with the following evaluation criteria and returns structured scores between 0 and 4, where higher values indicate better performance:

\begin{center}
\begin{graylist}
\begin{enumerate}[
    label=\textit{\arabic*)},
    leftmargin=1.5em,       
    rightmargin=0em,       
    itemsep=2pt, topsep=2pt 
]
    \item \textit{\textbf{Instruction Alignment (0-4):} Measures how well the video follows the described structure and sequence in the prompt. A high score indicates that the visual steps faithfully reflect the textual instructions.}
    \item \textit{\textbf{Temporal Consistency (0-4):} Evaluates the smoothness and continuity between frames. Disjointed or abrupt transitions will lead to a lower score.}
    \item \textit{\textbf{Visual Stability (0-4):} Assesses the stability of the video in terms of camera motion, object appearance, and scene composition. Shaky or glitchy outputs are penalized.}
    \item \textit{\textbf{Content Fidelity (0-4):} Determines how accurately the key elements described in the prompt are preserved. Hallucinated or missing objects/events will reduce the score.}
    \item \textit{\textbf{Focus Relevance (0-4):} Examines whether the video's visual attention remains focused on the correct objects or regions throughout. Irrelevant distractions or poorly framed targets are penalized.}
\end{enumerate}
\end{graylist}
\end{center}

We adopt a direct prompting strategy, instructing Gemini with the prompt, videos, and evaluation criteria to produce numerical scores in JSON format directly.

\begin{table*}[t]
\label{overall_result}
\centering
\small
\caption{\textbf{Model-level Overall and Per-dimension Performance on \textsc{MME-CoF}.} Mean scores and standard deviations are reported on a 0–4 scale, as graded by Gemini-2.5-Pro.}
\begin{adjustbox}{width=\textwidth}
\begin{tabular}{lccccccc}
\toprule
\makecell{Model} &
\makecell{Overall} &
\makecell{{Instruction}\\{Alignment}} &
\makecell{{Temporal}\\{Consistency}} &
\makecell{{Visual}\\{Stability}} &
\makecell{{Content}\\{Fidelity}} &
\makecell{{Focus}\\{Relevance}} \\
\cmidrule(lr){1-1} \cmidrule(lr){2-2} \cmidrule(lr){3-7}
Kling-v1~\cite{kuaishou2024kling} & 0.64~$\pm$~0.91 & 0.01~$\pm$~0.09 & 0.15~$\pm$~0.75 & \textbf{2.43~$\pm$~1.86} & 0.21~$\pm$~0.79 & 0.43~$\pm$~1.07 \\
Seedance-1.0-pro~\cite{gao2025seedance} & 1.41~$\pm$~1.51 & 0.30~$\pm$~0.86 & \textbf{1.65~$\pm$~1.57} & 2.00~$\pm$~1.72 & 1.13~$\pm$~1.65 & 1.98~$\pm$~1.75 \\
Veo-3.0-fast~\cite{GoogleDeepMind2025Veo3} & 1.44~$\pm$~1.51 & 0.56~$\pm$~1.09 & 1.37~$\pm$~1.51 & 1.88~$\pm$~1.73 & 1.10~$\pm$~1.52 & 2.27~$\pm$~1.69 \\
Veo-3.0-preview~\cite{GoogleDeepMind2025Veo3} & 1.45~$\pm$~1.50 & 0.54~$\pm$~1.06 & 1.43~$\pm$~1.53 & 1.89~$\pm$~1.71 & 1.12~$\pm$~1.49 & 2.26~$\pm$~1.73 \\
Sora-2-pro~\cite{openai2025sora2} &1.66~$\pm$~1.53 & 0.48~$\pm$~0.96 & 1.36~$\pm$~1.59 & 2.39~$\pm$~1.65 & 1.64~$\pm$~1.72 & 2.44~$\pm$~1.73   \\
Sora-2~\cite{openai2025sora2} & \textbf{1.72~$\pm$~1.59} & \textbf{0.59~$\pm$~1.12} & 1.52~$\pm$~1.69 & 2.32~$\pm$~1.68 & \textbf{1.62~$\pm$~1.75} & \textbf{2.52~$\pm$~1.71} \\
\bottomrule
\end{tabular}
\label{tab:model_dimension}
\end{adjustbox}
\vspace{0.2cm}
\end{table*}

\begin{table*}[t]
\centering
\small
\label{per_category_result}
\caption{\textbf{Per-category Scores on \textsc{MME-CoF}.} Mean scores and standard deviations are reported on a 0–4 scale, as graded by Gemini-2.5-Pro.}
\begin{adjustbox}{width=\textwidth}
\begin{tabular}{lcccccc}
\toprule
\makecell{Category} &
\makecell{Kling-v1~\cite{kuaishou2024kling}} &
\makecell{Seedance-1.0\\Pro~\cite{gao2025seedance}} &
\makecell{Veo-3.0\\Fast~\cite{GoogleDeepMind2025Veo3}} &
\makecell{Veo-3.0\\Preview~\cite{GoogleDeepMind2025Veo3}} &
\makecell{Sora-2~\cite{openai2025sora2}} &
\makecell{Sora-2\\Pro~\cite{openai2025sora2}} \\
\cmidrule(lr){1-1} \cmidrule(lr){2-2} \cmidrule(lr){3-3} \cmidrule(lr){4-4} \cmidrule(lr){5-5} \cmidrule(lr){6-6} \cmidrule(lr){7-7}
Visual Detail      & 0.72~$\pm$~0.69 & 1.37~$\pm$~1.39 & 1.10~$\pm$~1.24 & {1.59~$\pm$~1.68} & {1.14~$\pm$~1.32} & 1.08~$\pm$~1.89\\
Visual Trace       & 0.49~$\pm$~0.65 & 1.23~$\pm$~1.13 & 1.43~$\pm$~1.26 & 1.48~$\pm$~1.24 & {1.51~$\pm$~1.37} & {1.75~$\pm$~1.31}\\
Real-world Spatial & 0.77~$\pm$~0.76 & 1.79~$\pm$~1.53 & 2.07~$\pm$~1.54 & {2.10~$\pm$~1.46} & {1.84~$\pm$~1.43} & 1.77~$\pm$~1.35\\
3D Geometry        & 0.61~$\pm$~0.58 & {1.95~$\pm$~1.64} & 1.71~$\pm$~1.54 & 1.54~$\pm$~1.43 & {1.37~$\pm$~1.49} & 1.42~$\pm$~1.45\\
2D Geometry        & 0.49~$\pm$~0.67 & 0.96~$\pm$~1.11 & 1.18~$\pm$~1.15 & 1.27~$\pm$~1.20 & {1.77~$\pm$~1.45} & {1.77~$\pm$~1.21}\\
Physics-based      & 0.60~$\pm$~0.62 & 1.27~$\pm$~1.25 & 1.44~$\pm$~1.39 & 1.44~$\pm$~1.35 & {2.13~$\pm$~1.32} & 2.10~$\pm$~1.33\\
Rotation           & 0.22~$\pm$~0.34 & {2.30~$\pm$~1.46} & 1.83~$\pm$~1.44 & 1.60~$\pm$~1.29 & {1.62~$\pm$~1.37} & 1.44~$\pm$~1.28\\
Table \& Chart     & 0.87~$\pm$~0.72 & 0.71~$\pm$~1.18 & 0.82~$\pm$~1.30 & 0.96~$\pm$~1.44 & {1.84~$\pm$~1.61} & 1.48~$\pm$~1.59\\
GUI                & 1.09~$\pm$~0.51 & 0.70~$\pm$~0.76 & 1.11~$\pm$~1.09 & 1.18~$\pm$~0.89 & {1.88~$\pm$~1.64} & 1.52~$\pm$~1.48\\
Object Counting    & 0.64~$\pm$~0.58 & 1.15~$\pm$~0.97 & 2.03~$\pm$~1.42 & 1.84~$\pm$~1.42 & {2.06~$\pm$~1.48} & 1.86~$\pm$~1.41\\
Embodied           & 0.80~$\pm$~0.00 & {1.82~$\pm$~1.67} & 1.33~$\pm$~1.57 & 1.18~$\pm$~1.46 & {1.30~$\pm$~1.51} & 1.40~$\pm$~1.42\\
Medical            & 1.15~$\pm$~1.17 & 1.56~$\pm$~1.41 & 0.27~$\pm$~0.39 & 0.30~$\pm$~0.58 & {2.08~$\pm$~1.56} & 1.81~$\pm$~1.42\\
\bottomrule
\end{tabular}
\label{tab:model_category}
\end{adjustbox}
\end{table*}

\subsection{Quantitative Results and Analysis}
We report the quantitative scores of the five evaluated models across the five reasoning dimensions in Table~\ref{tab:model_dimension}, and provide detailed per-category results in Table~\ref{tab:model_category} and Figure~\ref{fig:overview:radar}.

Overall, most models exhibit limited reasoning capability across all tasks in \textsc{MME-CoF}, reflected by generally low scores. Among the five dimensions, \textit{Visual Stability} achieves the highest average, indicating that current video models can generate smooth and coherent sequences. 
Yet, their behavior remains largely at the level of pattern replay rather than genuine reasoning.

The Sora-2 series~\cite{openai2025sora2} shows relative advantages in physics-based, embodied, and medical reasoning, while the Veo-3.0 series~\cite{GoogleDeepMind2025Veo3} performs comparatively better in real-world spatial reasoning. 
Seedance-1.0-pro~\cite{gao2025seedance} demonstrates relative strength in rotation and 3D geometry reasoning. 
These trends suggest that different models specialize in distinct reasoning aspects. 
However, their mean scores remain below 2.0 out of 4, highlighting substantial room for improvement and pointing to opportunities for more targeted enhancement in future development.

\section{Related Work}

\paragraph{Video Models.}
Video models have been progressively evolving both in the fields of video understanding and generation. 
For video understanding methods, earlier approaches, such as MViT~\cite{fan2021multiscale}, Video Swin Transformer~\cite{liu2022video}, and VideoMAE~\cite{tong2022videomae}, aim to learn a robust representation that fosters downstream tasks. 
With the rise of LLMs, recent approaches encode videos as tokens and exploit the language backbone for captioning~\cite{tong2025g}, event localization~\cite{tian2018audio}, and high-level reasoning~\cite{hu2025video, zhao2025mmvu}.
Video generation models have also attracted much attention. Closed system, including OpenAI’s Sora~\cite{openai2024sora,openai2025sora2},
Runway’s Gen-3~\cite{Runway2024Gen3}, Pika Labs~\cite{PikaLabs2024Pika}, Luma AI~ \cite{LumaLabs2024DreamMachine}, and Google DeepMind’s Veo series~\cite{DeepMind2024Veo2, GoogleDeepMind2025Veo3}, have exhibited impressive results. 
However, they remain inaccessible due to their closed-source nature. Open-source alternatives have recently become available: Stable Video Diffusion~\cite{stable_video_diffusion} introduces efficient training strategies, Hunyan-Video~\cite{hunyuanvideo} proposes systematic scaling, and Wan-2.1~\cite{wan} presents an efficient 3D VAE with expanded pipelines.

\paragraph{Reasoning with Video.}
The advent of large reasoning models~\cite{guo2025seed1,tong2025delving,guo2025can,wei2022chain}, such as OpenAI o1~\cite{openai2024o1} and DeepSeek-R1~\cite{guo2025deepseek}, has spurred the development of video reasoning benchmarks.
Most current methods~\cite{feng2025videor1,li2025videochatr1,meng2025openo3} employ MLLMs specialized in video reasoning understanding.
For example, Video-R1~\cite{feng2025videor1} specifically targets temporal reasoning capabilities by introducing a temporal group relative policy optimization (GRPO) loss. 
VideoChat-R1~\cite{li2025videochatr1} focuses on spatio-temporal reasoning abilities by training with GRPO and rule-based rewards. 
A two-stage training strategy, combining SFT and RL, is used by VideoRFT~\cite{wang2025videorft}. 
When trained on vast collections of images and videos, this strategy boosts the model’s ability to handle QA tasks, whether in general contexts or reasoning-focused ones.
These methods primarily focus on enhancing specific types of question-answering or captioning tasks.
Concurrently, \cite{wiedemer2025video} demonstrates the large potential of video generative models in video reasoning. 
These models have implicitly acquired world knowledge throughdemonstrates impressive performance on various tasks, includinging and reasoning capability. Yet, this direction has rarely been explored and only experimented with in zero-shot settings.

\paragraph{Evaluation of Video Models as Zero Shot Learner.}
Recently, several works have been exploring the zero-shot capability of video generation models in various domains, including general-purpose vision understanding~\cite{wiedemer2025video,fu2024video}, medical imaging~\cite{lai2025video}, and world models~\cite{wang2025videoverse}. \cite{wiedemer2025video} conducts experiments on Veo 3 with a variety of vision tasks that have not been explicitly included during training. The video model showcases surprising performance on multiple tasks like object segmentation, image editing, and even maze solving. \cite{lai2025video} later adopts a similar paradigm to medical images understanding tasks and finds video generation models also show powerful capabilities, \eg delineation of anatomical structures in CT scans, medical image segmentation, and even forecasting of future 3D CT phases. Besides, \cite{wang2025videoverse} shows that video generation models could also understand complex temporal causality and world knowledge in the real world, thereby serving as a world model~\cite{agarwal2025cosmos, phyworld}.

\section{Conclusions and Insights}
\paragraph{Video models demonstrate an intuitive understanding of the simple visual world.}
Recent video models can generate high-fidelity videos with realistic motion dynamics, suggesting that they have internalized substantial visual and structural knowledge about the world. 
Through qualitative results from our empirical study and quantitative results from the \textsc{MME-CoF} benchmark, our work confirms that these models do exhibit intuitive yet local reasoning potential. This emergent behavior, which aligns with the ``Chain-of-Frame'' (CoF) mechanism, is revealed across several common success patterns. \textit{(i) Fine-grained Grounding.} Models demonstrate a capability for fine-grained attribute and spatial grounding, especially when targets are visually distinct, as presented in visual detail reasoning tasks. \textit{(ii) Short-horizon Trace Consistency.} In Visual Trace Reasoning tasks, models can maintain short-term consistency in visual traces.
\textit{(iii) Emergent Tool-Use Simulation.} An emergent ability to follow CoF instructions that mimic tool-use is presented, such as drawing lines in 2D geometry, highlighting targets in object counting, or controlling the camera in table and chart reasoning. \textit{(iv) Foundational Spatial and Geometric Grasp.} This includes single-step 3D geometry transformations, understanding basic real-world spatial layouts, finding coherent sequential paths, and handling small-angle Rotations. \textit{(v) Preliminary Real-world Interaction}. Models display a preliminary comprehension of real-world interaction, generating coherent manipulation paths in embodied reasoning.

\paragraph{Complex visual reasoning reveals fundamental limitations.}
However, visual reasoning demands more than these foundational skills. It tests a model's ability to maintain long-horizon logical consistency, adhere to abstract constraints, and understand functional principles. In these complex areas, our study reveals fundamental limitations and several common failure patterns. \textit{(i) Causal and Physical Logic.} This is evident in physics-based reasoning, where the model generates implausible motion that violates basic causal principles, and in visual trace reasoning, where the generated sequences break causal order with illogical steps.
\textit{(ii) Long-horizon and Rule-grounded Reasoning.} In visual trace reasoning, models fail to maintain state and adhere to task-specific rules over extended sequences.
\textit{(iii) Geometric and Spatial Logic.} Models fail at multi-step or complex transformations in 3D/2D geometry and real-world spatial tasks, often breaking constraints or prioritizing visual plausibility over correctness.
\textit{(iv) Functional and Interaction Logic.} They merely imitate GUI actions without grasping their purpose and lack the necessary planning and stability for reliable Embodied tasks, often resorting to workarounds.
\textit{(v) Perceptual Precision and Specialized Knowledge.} This weakness appears when models fail to identify small or indistinct targets in visual detail reasoning, distort data in table and chart tasks, and fail to process specialized medical imagery due to a lack of domain understanding.

\paragraph{Current video models are not yet ready as standalone zero-shot reasoners.}
Overall, our findings show that current video models are not yet reliable as standalone zero-shot reasoners. 
Strong generative performance does not automatically imply robust reasoning during inference. 
The model's behavior appears to be driven more by learning surface-level patterns and correlations rather than by internalizing general principles. It excels at short-term coherence rather than long-horizon causality. 
This is evident when the model prioritizes visual plausibility over precise spatial reasoning, or favors visually symmetric patterns over strictly adhering to geometric instructions. This tendency to produce plausible but instructionally flawed outputs reveals a reasoning process that is pattern-driven, not principle-driven, thereby undermining its ability to function as a standalone zero-shot reasoner.

\paragraph{The potential in advancing next-generation collaborative visual reasoning.}
Despite these limitations, the emergent behaviors observed in video models signal strong potential. The CoF concept suggests a novel modality for reasoning through visual problems step by step. While these models are not yet robust standalone reasoners, their foundational capabilities demonstrate that they can be guided through carefully designed prompts. 
This suggests a path where video models exhibit encouraging signs as complementary visual engines alongside dedicated reasoning models.
{
    \bibliographystyle{plain}
    \bibliography{main}
}
\end{document}